\documentclass[10pt,twocolumn,letterpaper]{article}

\usepackage{cvpr}              

\usepackage{threeparttable}
\usepackage{booktabs}
\usepackage[table]{xcolor}
\usepackage{threeparttable}

\definecolor{cvprblue}{rgb}{0.21,0.49,0.74}
\usepackage[pagebackref,breaklinks,colorlinks,allcolors=cvprblue]{hyperref}
\usepackage{subcaption}
\usepackage[table]{xcolor}
\usepackage{booktabs}
\usepackage{array}
\usepackage{tabularx}
\usepackage{enumitem}
\usepackage{makecell}
\usepackage{pifont}
\usepackage{booktabs}
\usepackage{subcaption}
\usepackage{graphicx}
\usepackage{pifont}
\usepackage{rotating}
\usepackage[table]{xcolor}
\usepackage{booktabs}
\usepackage{pifont}

\definecolor{lowdose}{RGB}{255,245,245}    
\definecolor{sparse}{RGB}{240,247,255}     
\definecolor{ours}{RGB}{240,255,240}       
\definecolor{mar}{RGB}{245,240,255}         
\definecolor{beam}{RGB}{255,245,230}       

\def\paperID{31} 
\def\confName{CVPR}
\def\confYear{2026}

\title{CT-DegradBench: A Physics-Informed Benchmark for CT Degradation Detection and Severity Estimation}

\author{
 Yousra Nabila Taifour$^{1}$, Marouane Tliba$^{1}$, Zuheng Ming$^{1}$, Marie Luong$^{1}$,\\
Nour Aburaed$^{2}$, Aladine Chetouani$^{1}$,
Gorkem Durak$^{3}$, Alessandro Bruno$^{4}$,\\
Faouzi Alaya Cheikh$^{5}$, Habib Zaidi$^{6}$, Ulas Bagci$^{3}$, Azeddine Beghdadi$^{1}$\\
$^{1}$Université Sorbonne Paris Nord, $^{2}$University of Dubai\\
$^{3}$Northwestern University, $^{4}$IULM University\\
$^{5}$Norwegian University of Science and Technology, $^{6}$University of Geneva\\
{\tt\small taifouryousra2017@gmail.com}
}
\usepackage[table]{xcolor}
\begin{document}
\maketitle
\begin{abstract}
Computed tomography (CT) images are frequently degraded by acquisition  artifacts, including noise, blur, streaking, aliasing, and metal artifacts. Yet CT enhancement is still largely evaluated using image quality metrics with limited perceptual and clinical validity, while existing datasets remain focused on isolated restoration tasks, hindering unified benchmarking across diverse degradation types.
We present \textbf{CT-DegradBench}, a dataset and benchmark for CT degradation detection and severity estimation under controlled single- and mixed-artifact settings. CT-DegradBench enables systematic evaluation across multiple degradation families and severity levels within a common experimental framework.
We further propose \textbf{SeSpeCT} (\textbf{Se}mantic--\textbf{Spe}ctral \textbf{CT} degradation estimation), a framework that combines semantic priors from medical vision--language models with complementary frequency-domain cues for artifact analysis. SeSpeCT constructs a training-free semantic quality axis in the multimodal embedding space using radiology-informed text prompts, without task-specific fine-tuning, and combines it with spectral features that capture degradation-specific frequency patterns. The resulting representation enables joint prediction of artifact type and severity. Experimental results show that SeSpeCT consistently outperforms the evaluated baselines under both single- and mixed-degradation settings.
The framework is available at \href{https://github.com/yousranb/CT-DEGRADBENCH}{https://github.com/yousranb/CT-DEGRADBENCH}.
\end{abstract}
\section{Introduction}
\label{sec:intro}



Computed tomography (CT) is a core medical imaging modality that provides high-resolution anatomical information essential for clinical diagnosis, treatment planning, and image-guided intervention \cite{mccollough2025advances,clement2025ai}. In routine practice, however, CT images are often degraded by artifacts introduced during acquisition and reconstruction, including noise, blur, streaking, aliasing, and metal artifacts \cite{clement2025ai,selles2024advances,yun2024tmaa,lei2023ct}. These degradations arise from diverse sources, such as dose reduction, sparse-view sampling, metallic implants, patient motion, and reconstruction parameter choices, and they frequently occur at different severity levels or in combination. Such effects compromise visual interpretation and negatively impact downstream automated analysis \cite{salimi2025deep}.

A large body of work has been devoted to CT image restoration, with deep learning methods now dominating low-dose denoising, sparse-view reconstruction, metal artifact reduction, and related enhancement tasks \cite{clement2025ai,lei2023ct,yun2024tmaa,sadia2024ct,dperceptct,tasharofi2025find,guan2023generative,wang2023review}. Yet despite this progress, evaluation still depends heavily on image quality assessment (IQA) metrics whose perceptual and clinical reliability across diverse degradation conditions remains limited \cite{Xun2025,kim2024systematic,clement2025ai,lei2023ct}. At the same time, most existing datasets are designed for isolated restoration settings, such as low-dose denoising, sparse-view reconstruction, or metal artifact reduction, rather than for unified analysis across multiple degradation types and severity levels. As a result, they provide limited support for systematic benchmarking of degradation detection, severity estimation, and IQA reliability under controlled conditions.

This exposes an important gap: while CT restoration has been widely studied, there is still no unified benchmark for analyzing how different degradations and severity levels affect CT image quality, nor for assessing whether commonly used metrics and modern representation models remain sensitive and reliable across such conditions. This problem is especially relevant in practice, where degradations rarely appear in a single canonical form and often co-occur in compound scenarios.


To address this gap, we introduce \textbf{CT-DegradBench}, a dataset and benchmark for CT artifact detection and severity estimation under controlled single- and mixed-degradation settings. CT-DegradBench is built from paired reference and degraded CT images generated using physics-informed degradation models and covers five common acquisition-related degradation types across calibrated severity levels. It also includes radiologist-informed mixed-artifact settings to approximate clinically plausible compound degradations.

\begin{figure*}[t!]
    \centering
    \includegraphics[width=\textwidth]{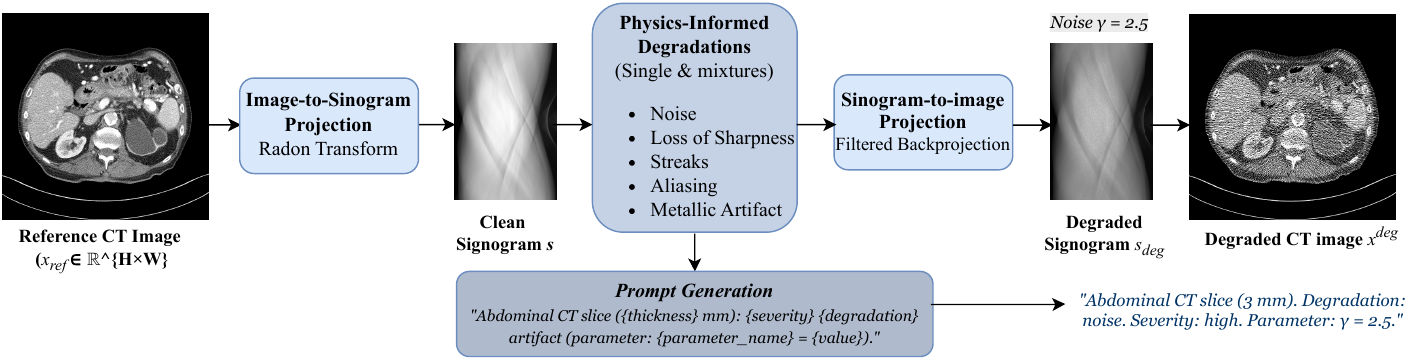}
    \caption{\textbf{CT-DegradBench generation pipeline.} A reference CT image is forward-projected to the sinogram domain, where physics-informed degradations are applied individually or as realistic mixtures with controlled severity. The degraded sinogram is reconstructed via filtered backprojection to obtain the final degraded CT image, while structured metadata are generated in parallel for prompt construction.}
    \label{fig:degradation_pipeline}
\end{figure*}
Using CT-DegradBench, we first conduct a systematic benchmark of classical and deep learning-based IQA metrics commonly used in CT enhancement evaluation, with a focus on degradation sensitivity and severity monotonicity. We then extend the benchmark to modern multimodal models by studying whether medical vision-language models (VLMs) encode quality-aware representations that correlate with degradation type and severity.

Building on this analysis, we propose \textbf{SeSpeCT} (\textbf{Se}mantic-\textbf{Spe}ctral \textbf{CT} degradation estimation), a framework for joint artifact detection and severity estimation. SeSpeCT constructs a training-free semantic quality axis in the multimodal embedding space of medical VLMs using radiology-informed text prompts, without task-specific fine-tuning, and combines this semantic signal with complementary spectral features that capture degradation-specific frequency structure. Together, these components provide a robust representation for CT degradation analysis under both isolated and mixed-artifact conditions. 
Our contributions are summarized as follows:
\begin{itemize}
    \item We introduce \textbf{CT-DegradBench}, a controlled benchmark for CT degradation detection and severity estimation, covering five common degradation types, calibrated severity levels, and radiologist-informed mixed-artifact settings.

    \item We provide the first \textbf{systematic evaluation of CT degradation sensitivity} across classical IQA metrics, learned perceptual metrics, and medical vision-language models, analyzing their behavior under diverse single- and mixed-artifact conditions.

    \item We propose \textbf{SeSpeCT}, a semantic-spectral framework that combines a training-free prompt-derived quality axis from medical vision-language models with frequency-domain features for joint artifact type and severity prediction.
\end{itemize}

\section{CT-DegradBench: Dataset and Benchmark}
\label{sec:methodology}
Existing CT datasets are largely designed for task-specific restoration problems, such as denoising or metal artifact reduction, and therefore do not support unified benchmarking across degradation types, severity levels, and mixed-artifact settings. We address this limitation with \textbf{CT-DegradBench}, a controlled multi-degradation CT dataset and benchmark for artifact detection and severity estimation. CT-DegradBench generates paired \textit{reference--degraded} CT slices with explicit degradation labels, calibrated severity levels, and clinically plausible artifact mixtures, enabling systematic evaluation under a common experimental framework.

\subsection{Problem setting and dataset overview}

Let $x^{\mathrm{ref}} \in \mathbb{R}^{H \times W}$ denote a reference CT image. We generate a degraded counterpart $x^{\mathrm{deg}}$ by applying one or more degradation operators $\mathcal{D}_k(\cdot)$ with known severity:
\begin{equation}
x^{\mathrm{deg}} = \mathcal{D}_{K}\circ \cdots \circ \mathcal{D}_{2}\circ \mathcal{D}_{1}\left(x^{\mathrm{ref}}\right),
\end{equation}
where each operator corresponds to a degradation type, namely noise, loss of sharpness, streaking, aliasing, or metal artifacts, and is associated with one of four severity levels (L0--L3), where L0 denotes the lowest severity level and L3 the highest. Each generated sample is accompanied by structured metadata specifying the degradation type(s), severity level(s), mixture composition, and generation parameters. This formulation enables controlled construction of both isolated and mixed degradations, making CT-DegradBench suitable for standardized benchmarking of degradation detection, severity estimation, and quality-aware representation learning.

\subsection{CT imaging background}

Computed tomography reconstructs cross-sectional images from X-ray projections acquired at multiple view angles. In clinical CT, reconstructed images are commonly represented in Hounsfield Units (HU), which encode attenuation relative to water. Given a reference CT slice $x^{\mathrm{ref}}$ in HU, we convert it to linear attenuation coefficients as
\begin{equation}
\mu = \mu_{\mathrm{w}}\left(1 + \frac{x^{\mathrm{ref}}}{1000}\right),
\end{equation}
where $\mu_{\mathrm{w}}$ denotes the attenuation coefficient of water \cite{hubbell_seltzer_2004_xray}.

The acquisition process is modeled in the projection domain. An ideal sinogram $s$ is obtained by applying the parallel-beam Radon transform
\begin{equation}
s = \mathcal{R}(\mu).
\end{equation}
Image reconstruction is then performed using filtered backprojection (FBP) \cite{brenner2007computed}:
\begin{equation}
\hat{\mu} = \mathcal{R}^{-1}_{\mathrm{FBP}}(s), \qquad
\hat{x} = 1000\left(\frac{\hat{\mu}}{\mu_{\mathrm{w}}}-1\right),
\label{FBP}
\end{equation}
where $\hat{x}$ is the reconstructed CT image in HU.

This formulation provides a physics-grounded basis for degradation simulation. In CT-DegradBench, degradations are introduced prior to reconstruction, either directly in the projection domain or by modifying the attenuation map in the image domain (e.g., for metal artifacts), and are subsequently propagated to the image domain through filtered backprojection.

\subsection{Degradation modeling and severity levels}

To enable controlled benchmarking, we model five common CT degradations, noise, loss of sharpness, streaking, aliasing, and metal artifacts, each at four calibrated severity levels (L0--L3).

\noindent\textbf{(1) Noise (mixed Poisson--Gaussian, severity $\gamma$).}
CT noise arises mainly from photon counting statistics and electronic detector noise in low-dose CT imaging. We simulate it in the projection domain using the mixed Poisson--Gaussian model of \cite{zeng2015simple}. Given a clean sinogram $s$, photon counts are sampled as
\begin{equation}
K \sim \mathrm{Poisson}\!\left(\alpha I_0 e^{-s}\right),
\end{equation}
where $I_0$ is the incident photon intensity and $\alpha$ controls the effective dose level. Detector noise is modeled by
\begin{equation}
K' = K + \mathcal{N}(0,\sigma^2),
\end{equation}
and the noisy sinogram is obtained via
\begin{equation}
s_{\mathrm{noisy}} = -\log\!\left(\frac{K' + \delta}{\alpha I_0}\right),
\end{equation}
where $\delta$ stabilizes the logarithm.

To generate controlled degradation levels for benchmarking, we introduce a residual noise scaling mechanism that enables calibrated control of noise severity while preserving noise characteristics consistent with the underlying CT acquisition physics.
Specifically, the residual noise is scaled as
\begin{equation}
s_{\mathrm{noisy}}^{\gamma} = s + \gamma\left(s_{\mathrm{noisy}} - s\right),
\qquad \gamma \in \{1,\ 2,\ 2.5,\ 4\}.
\end{equation}

The degraded sinogram is then reconstructed using FBP (Eq.~\ref{FBP}).

\noindent\textbf{(2) Loss of sharpness / blur (severity $\sigma$).}
The dominant resolution loss in CT arises from detector aperture and detector response, which primarily affect spatial resolution along the detector direction rather than across projection angles. Accordingly, we blur the sinogram along the detector axis with a 1D Gaussian kernel:
\begin{equation}
s_{\text{blur}} = G_{\sigma} * s, \quad \sigma \in \{0.8, 1, 1.5, 2.5\},
\end{equation}
where $*$ denotes convolution and $G_{\sigma}$ is applied along the detector dimension. The blurred sinogram is then reconstructed using FBP.

\noindent\textbf{(3) Streaks ( severity $\Delta L$).}
In practice, streak artifacts often co-occur with noise and other degradations. To isolate their effect, we introduce structured outliers in the sinogram that mimic directional streaks observed in clinical CT images:
\vspace{-2mm}
\begin{equation}
s_{\mathrm{streak}} = s + \Delta L \cdot M,
\qquad \Delta L \in \{0.25,\ 0.5,\ 1,\ 2\},
\end{equation}
\begin{figure*}[t]
    \centering  \resizebox{0.8\textwidth}{!}{
\includegraphics{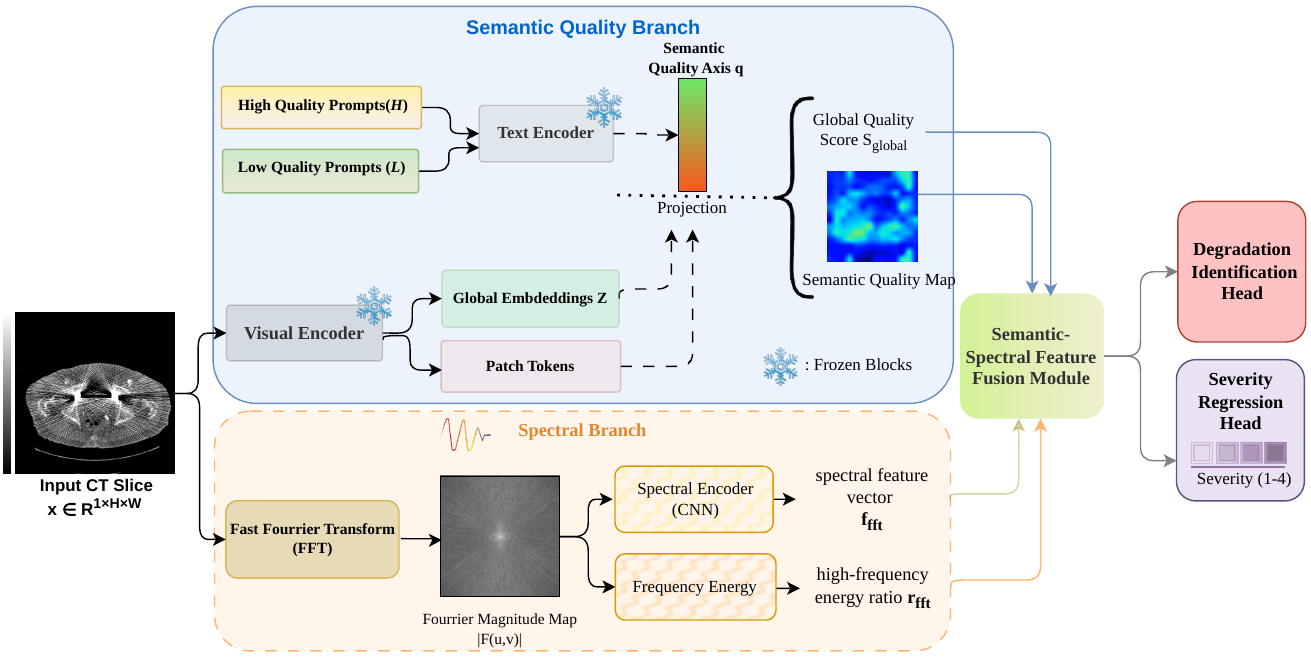} }\caption{\textbf{Overview of the proposed SeSpeCT framework.} 
The model combines a semantic quality branch derived from a medical vision–language model with frequency-domain descriptors extracted from the Fourier spectrum. 
The fused representation is used to jointly predict degradation type and severity.}
\label{fig:proposed_method}
\end{figure*}

\vspace{-1mm}
where $M$ is a binary mask selecting the affected regions. Reconstructing $s_{\mathrm{streak}}$ yields the streak-corrupted image.
\noindent\textbf{(4) Aliasing artifacts (severity: number of views).}
Aliasing artifacts from sparse-view CT \cite{guan2023generative} are simulated by subsampling projection angles. Let $\Theta$ denote the full set of acquisition angles and $\Theta_N \subset \Theta$ a subset of $N$ views. We form a sparse sinogram
\begin{equation}
s_{N} = \mathcal{R}_{\Theta_{N}}(\mu),
\qquad N \in \{180,\ 90,\ 60,\ 45\},
\end{equation}
and reconstruct it using FBP. Fewer views produce stronger aliasing artifacts.

\noindent\textbf{(5) Metal artifacts.}
Metal artifacts are simulated by inserting a metal region into the attenuation map, following \cite{haneda2025aapm}. Given a binary metal mask $m(\mathbf{x}) \in \{0,1\}$ and metal attenuation coefficient $\mu_{\mathrm{metal}}$, we define
\begin{equation}
\mu_{\mathrm{metal}}(\mathbf{x}) =
(1-m(\mathbf{x}))\,\mu(\mathbf{x}) + m(\mathbf{x})\,\mu_{\mathrm{metal}}.
\end{equation}
We then forward-project and reconstruct using Eq.~\ref{FBP}, yielding artifacts similar to those observed around metallic implants. Severity is controlled by increasing the size of the metal region. Because metal artifacts are spatially localized, we also provide bounding-box annotations derived from the mask $m(\mathbf{x})$.
\vspace{-5mm}
\paragraph{Mixed degradations.}
In clinical CT, multiple artifacts often co-occur due to interacting acquisition conditions, reconstruction processes, and patient-specific factors. To reflect this setting, CT-DegradBench includes mixed degradations in addition to isolated artifacts. We define five representative mixture configurations:
\textit{Blur + Noise},
\textit{Streaks + Noise},
\textit{Metal + Noise},
\textit{Aliasing + Noise}, and
\textit{Metal + Blur + Noise}.
These combinations capture common interactions between acquisition noise and other degradation mechanisms. Degradations are applied sequentially in an order that approximates their manifestation in reconstructed CT images; the corresponding clinical scenarios are summarized in the supplementary material (Table~\ref{tab:mixtures_scenarios}).

Each mixture is assigned a global severity level controlling the overall degradation strength. Component severities are sampled from neighboring levels around this global level to maintain comparable artifact magnitudes and avoid unrealistic combinations. We define the final mixture severity as the maximum severity among its constituent degradations.
\subsection{Metadata Description}
To support reproducible benchmarking, each CT-DegradBench sample is associated with structured metadata describing the degradation type(s), mixture order, severity level, and generation parameters. For metal artifacts, localization annotations are additionally provided. A structured natural-language prompt describing the degradation configuration is included to benchmark Vision–Language Models (VLMs), as illustrated in Figure~\ref{fig:degradation_pipeline}.
\vspace{-2mm}
\section{Proposed Method}
\label{proposed_method}

We propose \textbf{SeSpeCT}, a framework for CT degradation detection and severity estimation that combines semantic quality cues from a medical vision--language model with complementary Fourier-domain features. As shown in Fig.~\ref{fig:proposed_method}, the method consists of three parts: (1) a \textbf{Semantic Quality Branch} that derives quality-aware descriptors from a prompt-defined semantic axis, (2) a \textbf{Spectral Branch} that captures degradation-sensitive frequency patterns, and (3) a \textbf{fusion and prediction module} that jointly predicts degradation type and severity.
\subsection{Semantic Quality Branch}
\begin{figure}
 \centering
 \includegraphics[width=0.8\columnwidth]{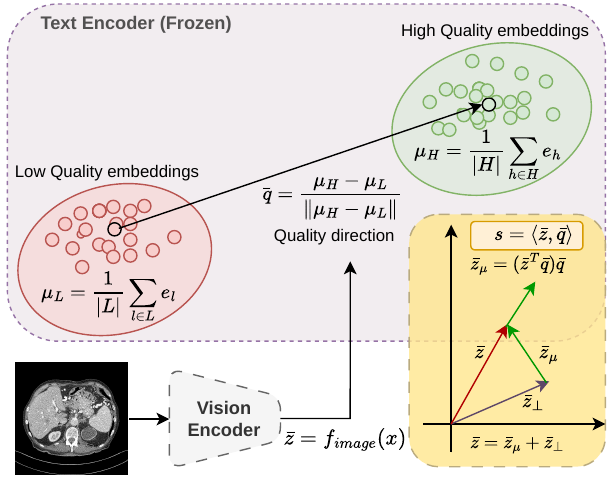}
 \caption{Semantic quality axis in the Vision-Language Model embedding space. High- and low-quality prompts produce embeddings $\bar{\mu}_H$ and $\bar{\mu}_L$, whose normalized difference defines the quality direction $\bar{q}$. Image embeddings $\bar{z}$ are projected onto this axis to obtain a quality score $s_{\text{global}} = \bar{z}^{\top}\bar{q}$.}
 \label{fig:semantic_quality_axis}
\end{figure}
\paragraph{Prompt-derived semantic quality axis.}
Let $\mathcal{H}$ and $\mathcal{L}$ denote two sets of radiology-informed prompts describing high-quality and degraded CT images, respectively. Using the frozen BioMedCLIP text encoder $f_{\text{text}}(\cdot)$ \cite{zhang2024bioMedCLIP}, we compute normalized prompt embeddings
\begin{equation}
e_h = \frac{f_{\text{text}}(h)}{\|f_{\text{text}}(h)\|}, \quad h \in \mathcal{H},
\qquad
e_l = \frac{f_{\text{text}}(l)}{\|f_{\text{text}}(l)\|}, \quad l \in \mathcal{L}.
\end{equation}
We then compute the corresponding prompt prototypes
\begin{equation}
\mu_H = \frac{1}{|\mathcal{H}|}\sum_{h \in \mathcal{H}} e_h,
\qquad
\mu_L = \frac{1}{|\mathcal{L}|}\sum_{l \in \mathcal{L}} e_l,
\end{equation}
and define the semantic quality axis as
\begin{equation}
q = \frac{\mu_H - \mu_L}{\|\mu_H - \mu_L\|}.
\end{equation}
As illustrated in Fig.~\ref{fig:semantic_quality_axis}, $q$ defines a direction in the shared vision--language embedding space that captures the transition from degraded to high-quality CT appearance.

\paragraph{Global and local semantic descriptors.}
Given an input CT slice $x$, the frozen BioMedCLIP image encoder produces a normalized global embedding
\begin{equation}
z = \frac{f_{\text{image}}(x)}{\|f_{\text{image}}(x)\|}.
\end{equation}
Projecting $z$ onto the semantic quality axis yields a global semantic quality score
\begin{equation}
s_{\text{global}} = z^\top q.
\end{equation}
To capture localized degradations, we additionally project Vision Transformer patch embeddings $T=\{t_1,\dots,t_N\}$ onto the same axis:
\begin{equation}
s_i = \hat{t}_i^\top q,
\qquad
\hat{t}_i = \frac{t_i}{\|t_i\|}.
\end{equation}
These scores form a semantic quality map over the image, as illustrated in Fig.~\ref{fig:proposed_method}. We summarize them using a pooling operator $\phi(\cdot)$ to obtain a fixed-length local descriptor
$
f_{\text{loc}} = \phi(s_1,\dots,s_N).
$
The final semantic representation is
$
f_{\text{sem}} = [\, s_{\text{global}} \Vert f_{\text{loc}} \,].
$
In practice, $\phi(\cdot)$ is implemented using simple statistics over patch responses, such as their mean, maximum, and standard deviation.
\subsection{Spectral Branch}

\begin{figure*}[t]
\centering

\begin{tabular}{cccccc}

\textbf{Reference} &
\textbf{Noise} &
\textbf{Loss of sharpness} &
\textbf{Streak artifacts} &
\textbf{Aliasing artifact} &
\textbf{Metal artifact} \\

\includegraphics[width=0.14\textwidth]{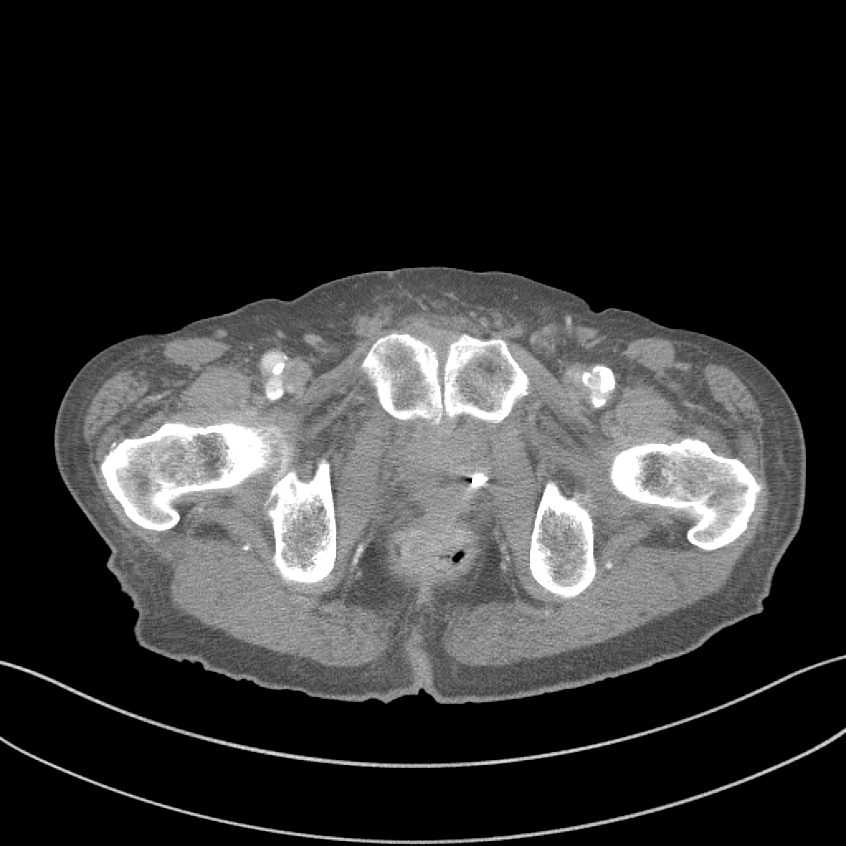} &
\includegraphics[width=0.14\textwidth]{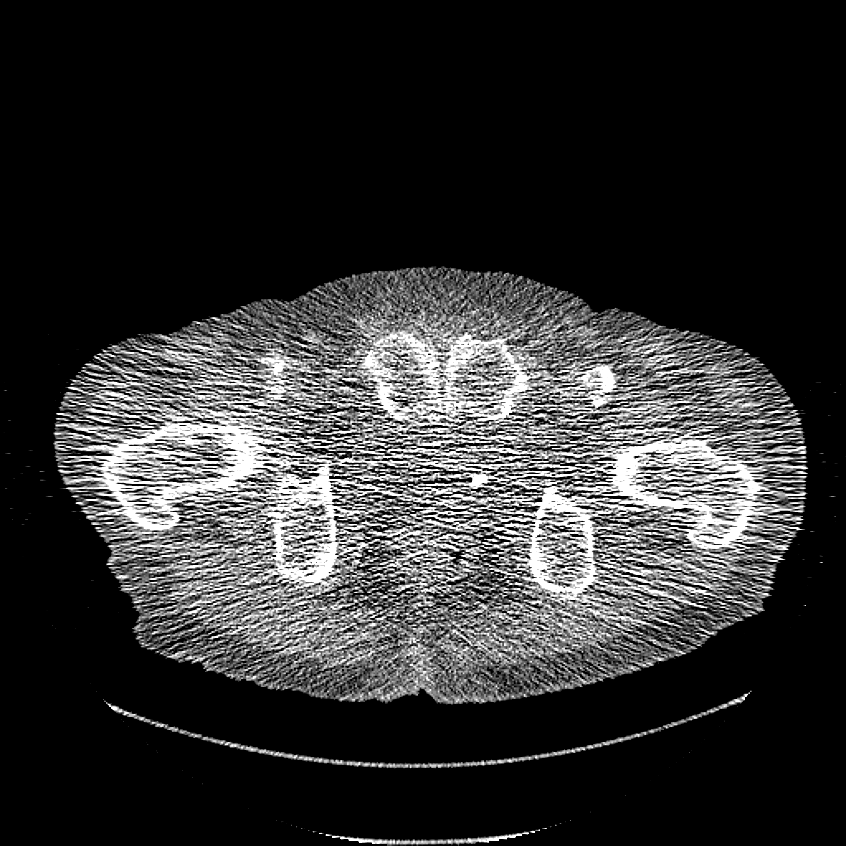} &
\includegraphics[width=0.14\textwidth]{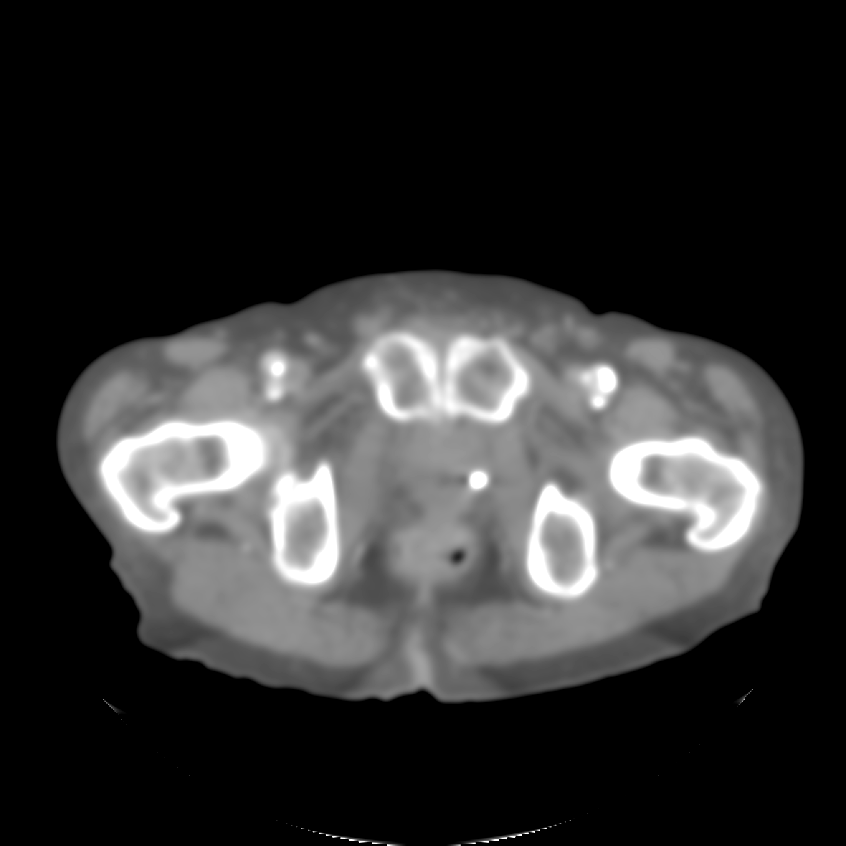} &
\includegraphics[width=0.14\textwidth]{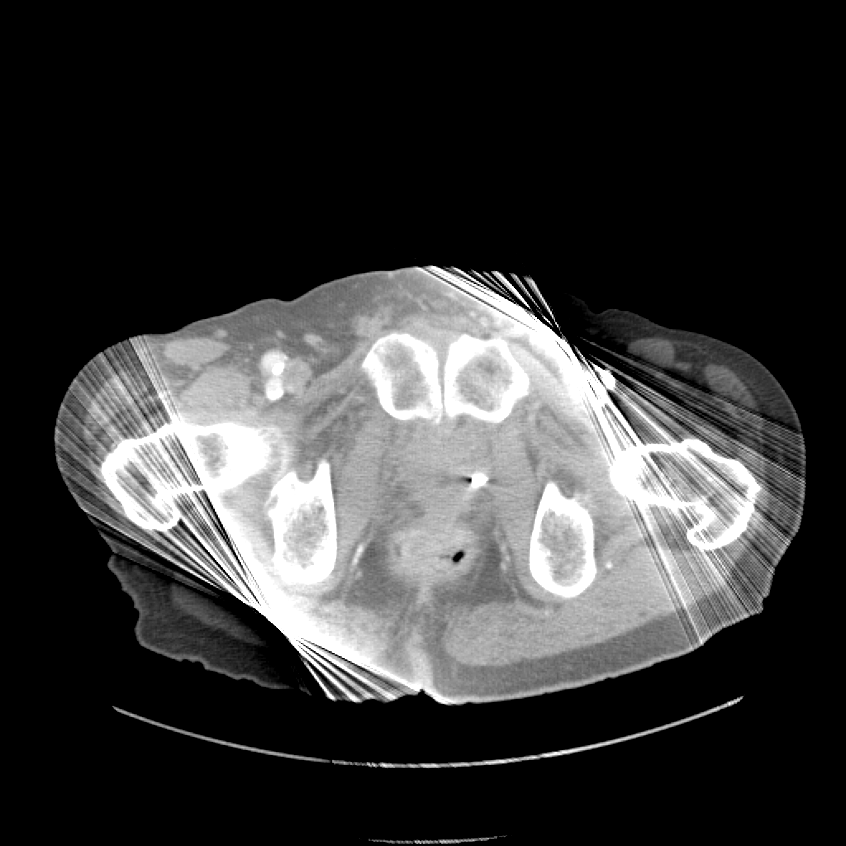} &
\includegraphics[width=0.14\textwidth]{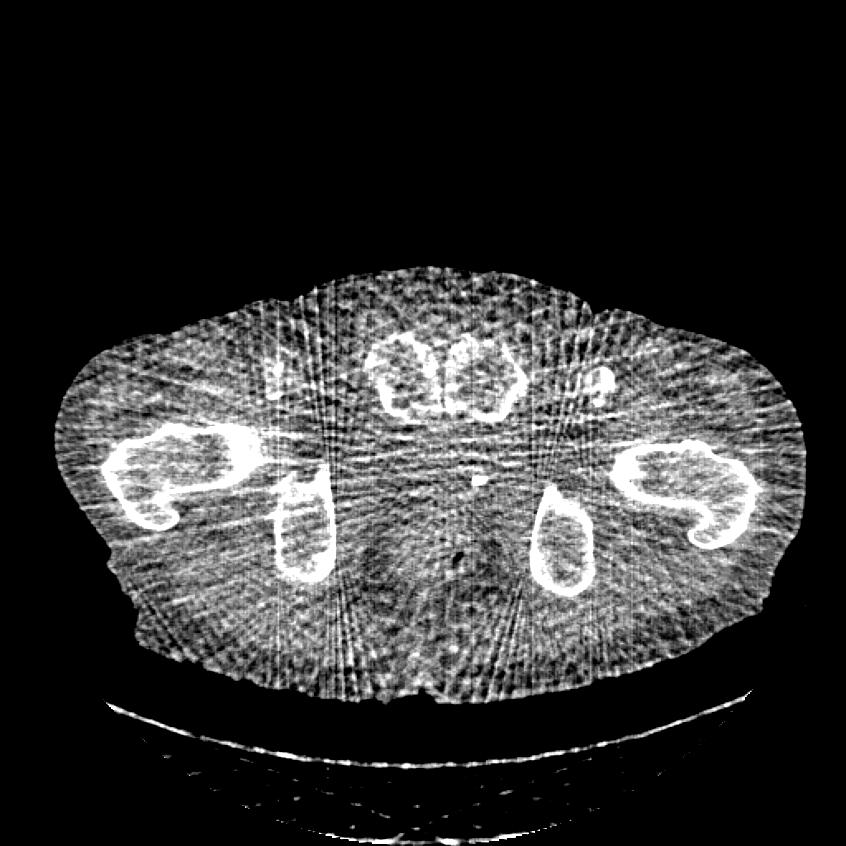} &
\includegraphics[width=0.14\textwidth]{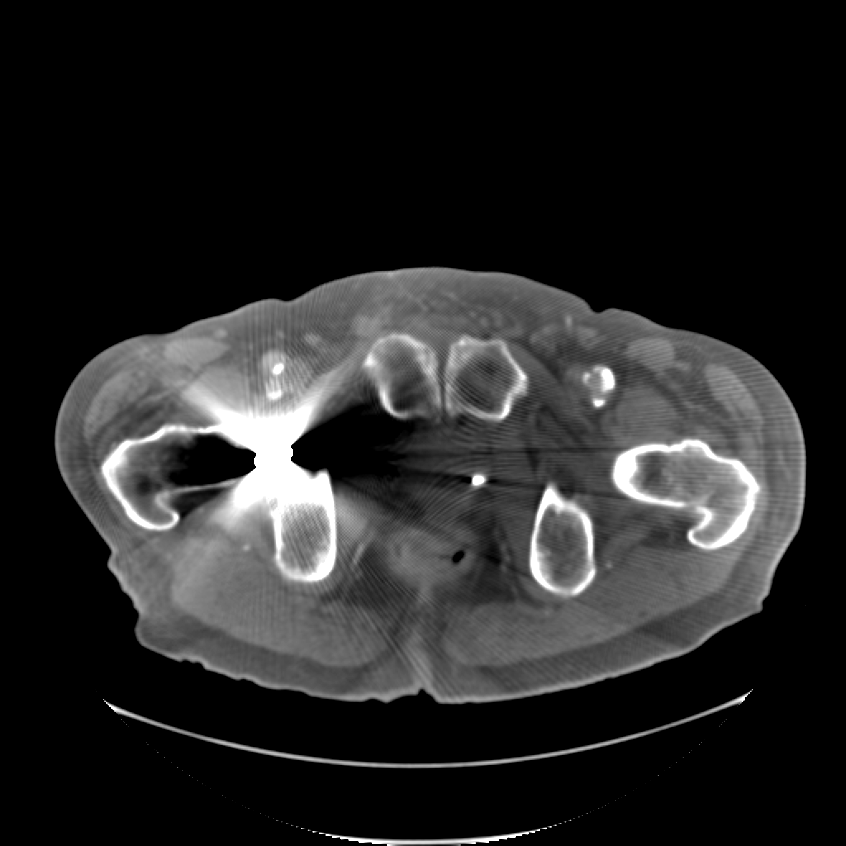} \\

\includegraphics[width=0.14\textwidth]{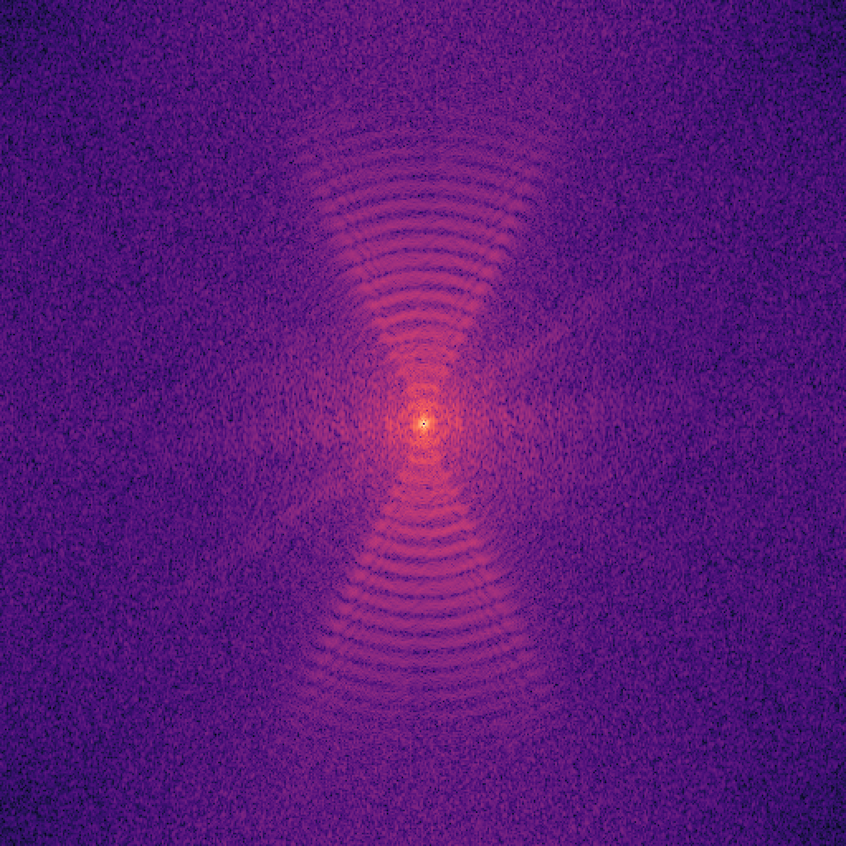} &
\includegraphics[width=0.14\textwidth]{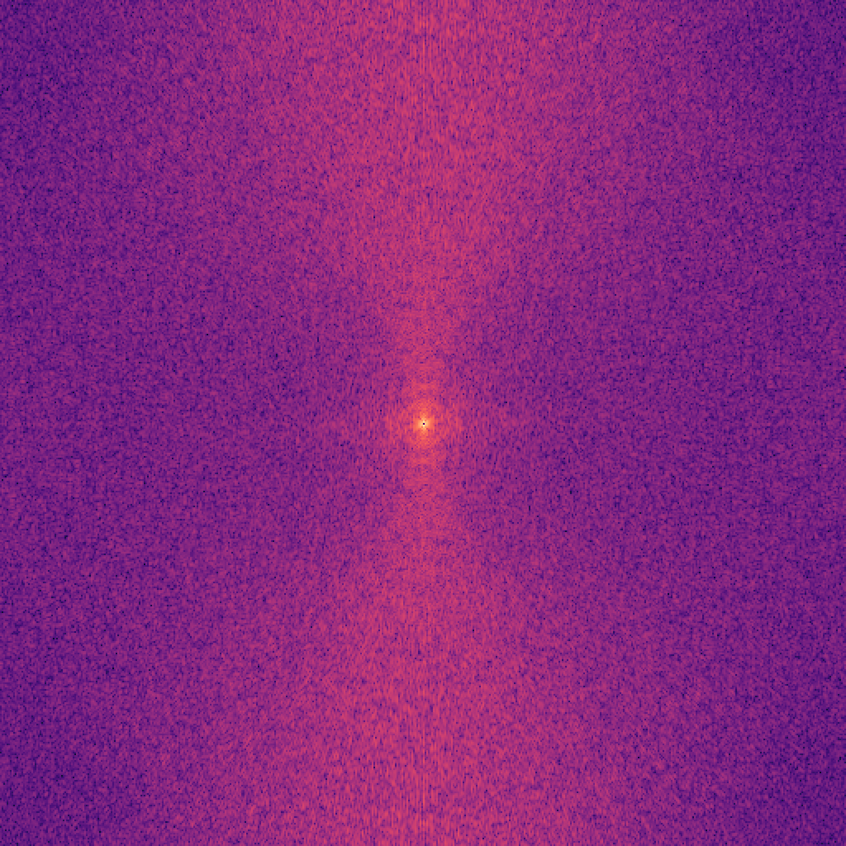} &
\includegraphics[width=0.14\textwidth]{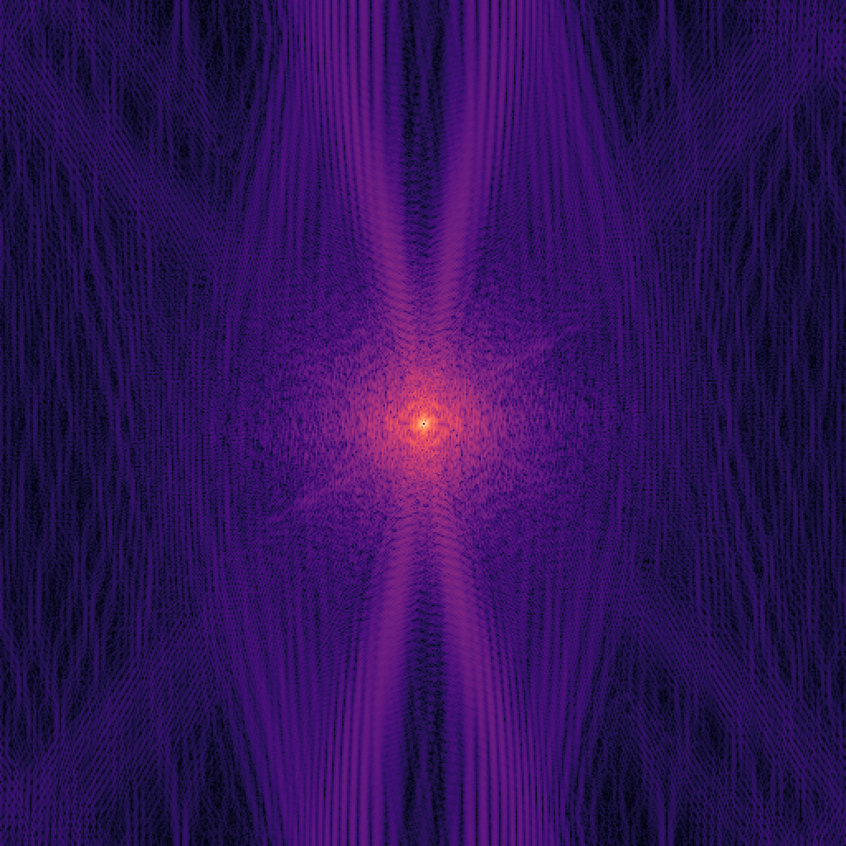} &
\includegraphics[width=0.14\textwidth]{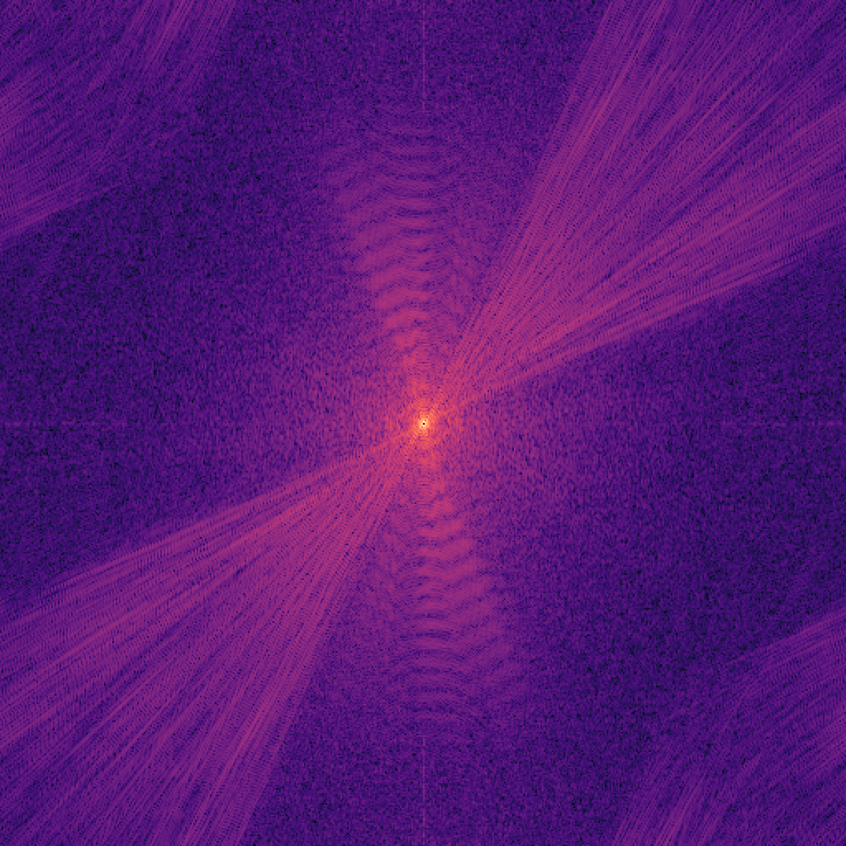} &
\includegraphics[width=0.14\textwidth]{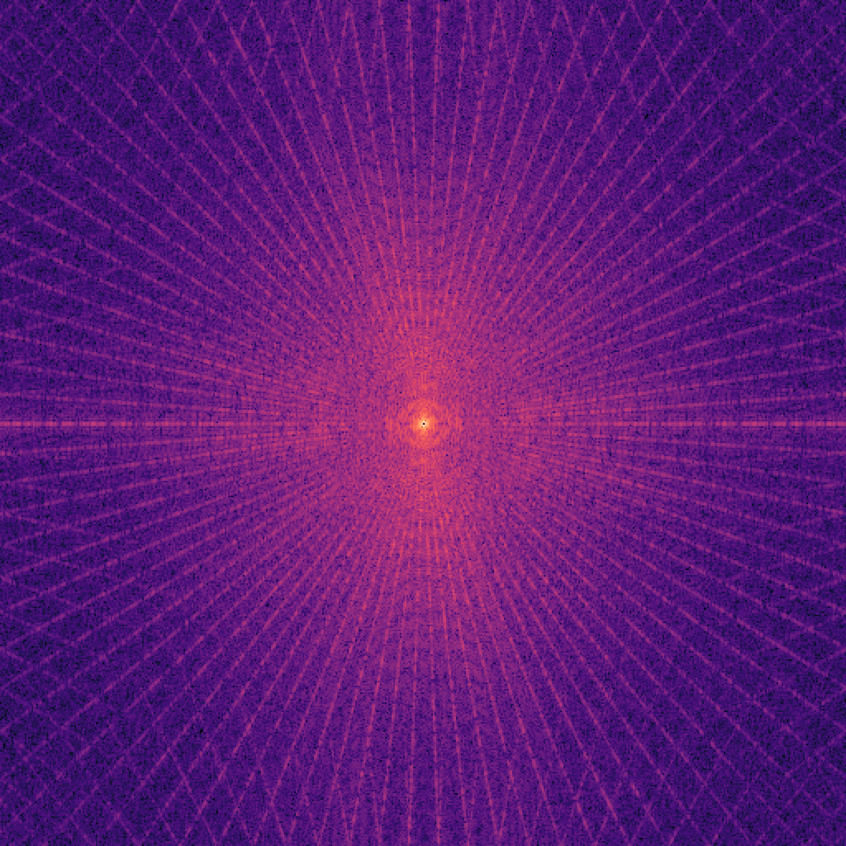} &
\includegraphics[width=0.14\textwidth]{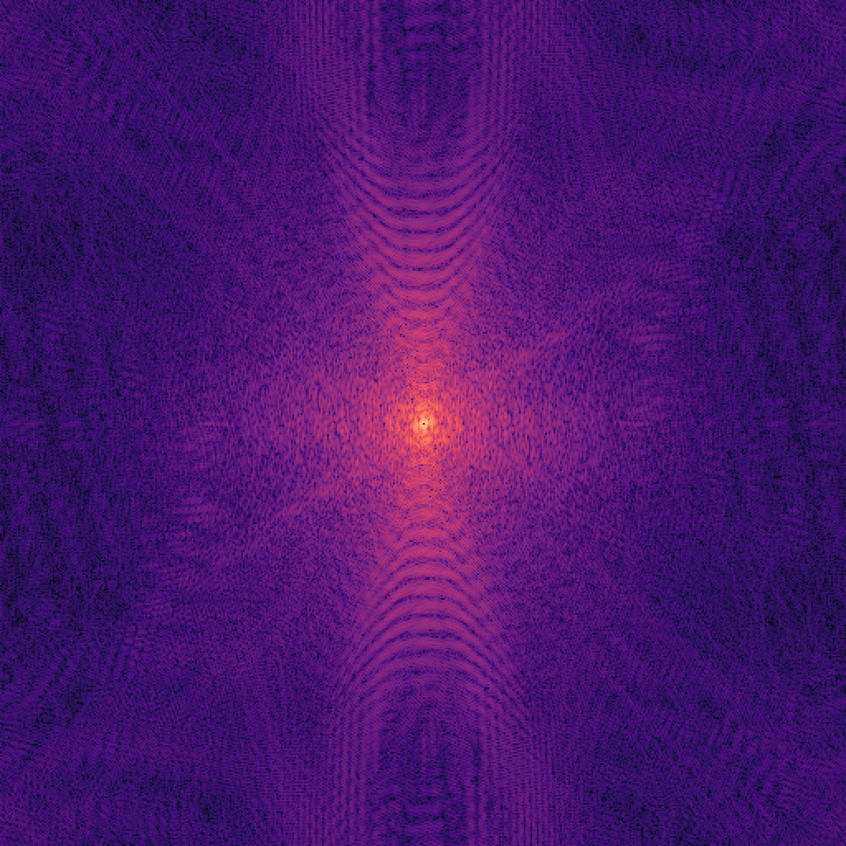} \\

\end{tabular}
\caption{Visualization of mid-severity degradations in the image domain (top) and Fourier domain (bottom). The distinct spectral patterns motivate the use of frequency-domain features as complementary cues for degradation detection and severity estimation.}
\label{fig:degradation_patterns_fourier}

\end{figure*}

Several CT degradations are more clearly expressed in the frequency domain than in the image domain. In particular, aliasing and streak artifacts produce structured spectral distortions, while noise alters the high-frequency energy distribution (See Fig. \ref{fig:degradation_patterns_fourier}). We therefore complement the semantic branch with frequency-domain features.

Given an input image $x$, we compute its discrete Fourier transform $F(u,v)=\mathcal{F}(x)(u,v)$ and form the log-magnitude spectrum $M(u,v)=\log(1+|F(u,v)|)$. This spectrum is then processed by a lightweight convolutional encoder $g_{\text{fft}}(\cdot)$ to obtain a spectral feature vector $h_{\text{fft}}$.

To capture the relative concentration of high-frequency content, we also compute a high-frequency energy ratio
\begin{equation}
r_{\text{fft}}=
\frac{\sum_{(u,v)\in\Omega_h}|F(u,v)|^2}
{\sum_{u,v}|F(u,v)|^2},
\end{equation}
where $\Omega_h$ denotes a predefined high-frequency region. The final spectral descriptor is formed by concatenating the learned spectral feature and the energy ratio, i.e., $f_{\text{spec}} = [\, h_{\text{fft}} \Vert r_{\text{fft}} \,]$.



\subsection{Fusion and Multi-Task Prediction}

The semantic descriptor $f_{\text{sem}}$ and spectral descriptor $f_{\text{spec}}$ provide complementary information: the former captures quality-aware semantic structure, while the latter captures degradation-specific frequency patterns. We concatenate them and project them into a shared embedding space using a multilayer perceptron:
\begin{equation}
z_f = \mathrm{MLP}\bigl([\, f_{\text{sem}} \Vert f_{\text{spec}} \,]\bigr),
\qquad z_f \in \mathbb{R}^{d}.
\end{equation}

From the fused embedding $z_f$, we predict both degradation category and severity. The classification head outputs
\begin{equation}
o = W_{\text{cls}} z_f + b_{\text{cls}},
\end{equation}
where $o \in \mathbb{R}^{C}$ are the logits over $C$ degradation classes. The predicted class is
\begin{equation}
\hat{y} = \arg\max_c \, \mathrm{softmax}(o)_c.
\end{equation}

The severity head predicts a scalar severity score
\begin{equation}
\hat{s} = w_{\text{reg}}^\top z_f + b_{\text{reg}}.
\end{equation}

\vspace{-1mm}
\subsection{Training Objective}

We train the model with a multi-task objective that combines degradation classification, severity regression, ordinal ranking, and supervised contrastive learning:
\begin{equation}
\mathcal{L}
=
\mathcal{L}_{\text{cls}}
+
\lambda_{\text{reg}}\mathcal{L}_{\text{reg}}
+
\lambda_{\text{rank}}\mathcal{L}_{\text{rank}}
+
\lambda_{\text{con}}\mathcal{L}_{\text{con}}.
\end{equation}
Here, $\mathcal{L}_{\text{cls}}$ is the standard cross-entropy loss for degradation classification, and $\mathcal{L}_{\text{reg}}=\mathrm{SmoothL1}(\hat{s},s)$ is the regression loss for severity prediction, where $s$ denotes the ground-truth severity label.

To preserve the ordinal structure of severity levels, we introduce a pairwise ranking loss over sample pairs $(i,j)$ such that $s_i > s_j$:
\begin{equation}
\mathcal{L}_{\text{rank}}
=
\frac{1}{|\mathcal{P}|}
\sum_{(i,j)\in\mathcal{P}}
\max\!\bigl(0, m - (\hat{s}_i - \hat{s}_j)\bigr),
\mathcal{P}=\{(i,j)\mid s_i>s_j\}.
\end{equation}

To further structure the fused embedding space, we apply a supervised contrastive loss on $z_f$, where samples sharing the same degradation class and severity level are treated as positives:
\begin{equation}
\mathcal{L}_{\text{con}}
=
-\frac{1}{|\mathcal{V}|}
\sum_{i \in \mathcal{V}}
\frac{1}{|P(i)|}
\sum_{p \in P(i)}
\log
\frac{
\exp(z_{f,i}^\top z_{f,p} / \tau)
}{
\sum_{a \neq i}\exp(z_{f,i}^\top z_{f,a} / \tau)
}.
\end{equation}

We set $\lambda_{\text{rank}}=0.3$ and $\lambda_{\text{con}}=0.05$. The classification and regression terms provide the main supervision, while the ranking and contrastive terms improve ordinal consistency and embedding structure.

\begin{table*}[t]
\centering

\scriptsize
\setlength{\tabcolsep}{3pt}
\renewcommand{\arraystretch}{1.05}

\begin{threeparttable}
\resizebox{\textwidth}{!}{%
\begin{tabular}{lccccc @{\hspace{0.6em}\vrule\hspace{0.6em}} lccccc}
\toprule
\rowcolor{gray!12}
\multicolumn{6}{c}{\textbf{Single distortions} ($\rho$ / $r$)} &
\multicolumn{6}{c}{\textbf{Mixtures} ($\rho$ / $r$)} \\
\cmidrule(lr){1-6}\cmidrule(lr){7-12}
\rowcolor{gray!12}
\textbf{Setting} & \textbf{PSNR} $\uparrow$ & \textbf{SSIM} $\uparrow$ & \textbf{VIF} $\uparrow$ & \textbf{LPIPS} $\downarrow$ & \textbf{DISTS} $\downarrow$ &
\textbf{Setting} & \textbf{PSNR} $\uparrow$ & \textbf{SSIM} $\uparrow$ & \textbf{VIF} $\uparrow$ & \textbf{LPIPS} $\downarrow$ & \textbf{DISTS} $\downarrow$ \\
\midrule

S1\_noise
& -0.6636/-0.6691 & -0.7095/-0.7087 & -0.6712/-0.6396 &  0.6463/0.6475 &  0.6979/0.6756 &
M1\_b+n
& -0.4894/-0.4924 & -0.5429/-0.5420 & -0.6273/-0.6655 &  0.5330/0.5360 &  0.5670/0.5940 \\

S2\_blur
& -0.2837/-0.2490 & -0.7936/-0.7747 & -0.8926/-0.8953 &  0.8301/0.8231 &  0.9107/0.9181 &
M2\_s+n
& -0.6129/-0.6150 & -0.6284/-0.6283 & -0.5475/-0.5781 &  0.5533/0.5544 &  0.5539/0.5743 \\

S3\_streak
& -0.6883/-0.6729 & -0.9345/-0.8719 & -0.7093/-0.6921 &  0.9447/0.9405 &  0.9068/0.8984 &
M3\_m+n
& -0.8055/-0.8024 & -0.8074/-0.8114 & -0.7244/-0.7641 &  0.7001/0.7191 &  0.7250/0.7666 \\

S4\_aliasing
& -0.5356/-0.5272 & -0.9518/-0.9521 & -0.9548/-0.9379 &  0.9600/0.9495 &  0.9548/0.9303 &
M4\_a+n
& -0.5437/-0.5499 & -0.6877/-0.6834 & -0.6203/-0.6583 &  0.6052/0.6102 &  0.5986/0.6209 \\

S5\_metal
& -0.6413/-0.6340 & -0.5317/-0.4794 & -0.2461/-0.2490 &  0.0431/0.0497 &  0.2929/0.2861 &
M5\_m+b+n
& -0.8062/-0.7941 & -0.8139/-0.8127 & -0.7941/-0.8246 &  0.7478/0.7620 &  0.7702/0.8043 \\
\midrule
\textbf{Mean}
& -0.5625/-0.5504 & \textbf{-0.7842/-0.7574} & -0.6948/-0.6828 & 0.6848/0.6821 & 0.7526/0.7417
&
\textbf{Mean}
& -0.6515/-0.6508 & \textbf{-0.6961/-0.6956} & -0.6627/-0.6981 & 0.6279/0.6363 & 0.6429/0.6720 \\

\bottomrule
\end{tabular}%
}

\begin{tablenotes}[flushleft]
\footnotesize
\item Entries are reported as \textbf{$\rho$/$r$} (Spearman/Pearson). Positive values indicate the metric increases with severity; negative values indicate it decreases with severity.
\item \textbf{Setting keywords:} S1-S5 = single degradations; M1-M5 = mixtures; L0-L3 = severity levels (increasing with $L$).
\item \textbf{Mixture shorthand:} b = blur, n = noise, s = streaks, a = aliasing, m = metal.
\end{tablenotes}
\end{threeparttable}
\caption{Spearman rank correlation ($\rho$) and Pearson correlation ($r$) results between objective quality metrics and degradation severity levels. (Full Dataset)}
\label{tab:corr_metrics_degradations_full}
\end{table*}

\begin{table*}[t]
\centering

\scriptsize
\setlength{\tabcolsep}{3pt}
\renewcommand{\arraystretch}{1.05}

\begin{threeparttable}
\resizebox{\textwidth}{!}{%
\begin{tabular}{lccccc @{\hspace{0.6em}\vrule\hspace{0.6em}} lccccc}
\toprule
\rowcolor{gray!12}
\multicolumn{6}{c}{\textbf{Single distortions} ($\rho$ / $r$)} &
\multicolumn{6}{c}{\textbf{Mixtures} ($\rho$ / $r$)} \\
\cmidrule(lr){1-6}\cmidrule(lr){7-12}
\rowcolor{gray!12}
\textbf{Setting} & \textbf{PSNR} $\uparrow$ & \textbf{SSIM} $\uparrow$ & \textbf{VIF} $\uparrow$ & \textbf{LPIPS} $\downarrow$ & \textbf{DISTS} $\downarrow$ &
\textbf{Setting} & \textbf{PSNR} $\uparrow$ & \textbf{SSIM} $\uparrow$ & \textbf{VIF} $\uparrow$ & \textbf{LPIPS} $\downarrow$ & \textbf{DISTS} $\downarrow$ \\
\midrule

S1\_noise
& -0.6644/-0.6743 & -0.7470/-0.7411 & -0.6873/-0.6644 &  0.7120/0.7175 &  0.7321/0.7255 &
M1\_b+n
& -0.5221/-0.5254 & -0.6095/-0.5978 & -0.7084/-0.6673 &  0.6319/0.6169 &  0.6726/0.6362 \\

S2\_blur
& -0.3108/-0.2307 & -0.8631/-0.8600 & -0.9126/-0.9209 &  0.8335/0.8316 &  0.9144/0.9202 &
M2\_s+n
& -0.6245/-0.6287 & -0.6538/-0.6613 & -0.6057/-0.5824 &  0.6095/0.6129 &  0.5982/0.5916 \\

S3\_streak
& -0.7252/-0.6910 & -0.9449/-0.8697 & -0.7453/-0.7256 &  0.9427/0.9366 &  0.9271/0.9182 &
M3\_m+n
& -0.8035/-0.8054 & -0.8258/-0.8300 & -0.7592/-0.7223 &  0.7661/0.7518 &  0.7656/0.7376 \\

S4\_aliasing
& -0.4856/-0.5009 & -0.9432/-0.9386 & -0.9660/-0.9541 &  0.9632/0.9501 &  0.9415/0.9209 &
M4\_a+n
& -0.5438/-0.5497 & -0.7243/-0.7093 & -0.6638/-0.6343 &  0.6852/0.6774 &  0.6441/0.6267 \\

S5\_metal
& -0.6261/-0.6401 & -0.4833/-0.4559 & -0.2847/-0.2660 &  0.0460/0.0489 &  0.3169/0.3318 &
M5\_m+b+n
& -0.8161/-0.8058 & -0.8362/-0.8378 & -0.8282/-0.7998 &  0.8136/0.7955 &  0.8362/0.7993 \\

\midrule
\textbf{Mean}
& -0.5624/-0.5474 
& \textbf{-0.7963/-0.7731 }
& -0.7192/-0.7062 
& 0.6995/0.6969 
& 0.7664/0.7633
&
\textbf{Mean}
& -0.6628/-0.6630 
& \textbf{-0.7299/-0.7272} 
& -0.7131/-0.6812 
& 0.7013/0.6909 
& 0.7033/0.6783 \\

\bottomrule
\end{tabular}%
}

\begin{tablenotes}[flushleft]
\footnotesize
\item Entries are reported as \textbf{$\rho$/$r$} (Spearman/Pearson). Positive values indicate the metric increases with severity; negative values indicate it decreases with severity.
\item \textbf{Setting keywords:} S1-S5 = single degradations; M1-M5 = mixtures.
\end{tablenotes}
\end{threeparttable}

\caption{Spearman rank correlation ($\rho$) and Pearson correlation ($r$) results between objective quality metrics and degradation severity levels (Test-Set; Patients: L506, L192, L310).
}
\label{tab:corr_metrics_degradations_testSet}
\end{table*}

\section{Experiments and Results}
\label{results}

\subsection{Experimental Setup}

\paragraph{Dataset and protocol.}
We construct CT-DegradBench from the Mayo 2016 Low-Dose CT Grand Challenge dataset \cite{mccollough2016tu}, which provides high-quality clinical CT scans together with acquisition parameters required for physics-informed degradation simulation. We select the high-dose CT acquisitions (200 mAs, 3 mm slice thickness, sharp reconstruction kernel) as reference images and generate degraded counterparts using the degradation pipeline described in Section~\ref{sec:methodology}. In total, the benchmark contains 2,378 reference slices of size $512\times512$ from 10 patients, from which we generate 9,512 degraded samples spanning multiple degradation types and severity levels. 

Unless otherwise stated, benchmark analyses of IQA metrics and frozen VLM representations are conducted on the full dataset to study their behavior across degradation types and severity levels. For evaluation of the proposed method, we additionally report results on a held-out test split comprising 30\% of the dataset in order to assess generalization to unseen samples.

\begin{table*}[h!]
\centering
\tiny
\setlength{\tabcolsep}{6pt}
\renewcommand{\arraystretch}{1}

\begin{threeparttable}
\begin{tabular}{lccccc lccccc}
\toprule
\rowcolor{gray!12}
\multicolumn{6}{c}{\textbf{Single distortions} ($\rho$ / $r$)} &
\multicolumn{6}{c}{\textbf{Mixtures} ($\rho$ / $r$)} \\
\cmidrule(lr){1-6} \cmidrule(lr){7-12}

\rowcolor{gray!12}
\textbf{Setting} &
\textbf{OpenCLIP \cite{cherti2023reproducible}} &
\textbf{MedCLIP \cite{wang2022MedCLIP}} &
\textbf{BioMedCLIP \cite{zhang2024bioMedCLIP}} &
\textbf{Merlin \cite{blankemeier2024merlin}} &
\textbf{Ours} &
\textbf{Setting} &
\textbf{OpenCLIP \cite{cherti2023reproducible}} &
\textbf{MedCLIP \cite{wang2022MedCLIP}} &
\textbf{BioMedCLIP \cite{zhang2024bioMedCLIP}} &
\textbf{Merlin \cite{blankemeier2024merlin}} &
\textbf{Ours} \\
\midrule
S1\_noise   & 0.611/0.538 & 0.345/0.276 & 0.682/0.582 & 0.621/0.484 & \textbf{0.798/0.791} &
M1\_b+n     & 0.543/0.493 & 0.236/0.216 & 0.663/0.582 & 0.437/0.291 & \textbf{0.709/0.685} \\

S2\_blur    & 0.649/0.596 & 0.533/0.486 & 0.885/0.796 & 0.169/0.191 & \textbf{0.887/0.869} &
M2\_s+n     & 0.485/0.420 & 0.301/0.283 & 0.657/0.615 & 0.495/0.363 & \textbf{0.668/0.657} \\

S3\_streak  & 0.769/0.617 & 0.595/0.551 & 0.893/0.858 & 0.399/0.418 & \textbf{0.926/0.894} &
M3\_m+n     & 0.653/0.655 & 0.228/0.215 & 0.542/0.518 & 0.679/0.605 & \textbf{0.745/0.737} \\

S4\_aliasing& 0.839/0.759 & 0.738/0.640 & 0.912/0.855 & 0.604/0.543 & \textbf{0.854/0.846} &
M4\_a+n     & 0.595/0.551 & 0.255/0.206 & 0.692/0.600 & 0.438/0.361 & \textbf{0.668/0.658} \\

S5\_metal   & 0.114/0.073 & 0.037/0.041 & 0.114/0.108 & 0.107/0.090 & \textbf{0.435/0.422} &
M5\_m+b+n   & 0.691/0.697 & 0.227/0.217 & 0.606/0.585 & 0.677/0.596 & \textbf{0.758/0.756} \\

\midrule
\rowcolor{gray!15}
\textbf{Mean} 
& 0.635/0.588
& 0.330/0.313
& 0.665/0.610
& 0.452/0.394
& \textbf{0.780/0.764}
&
\textbf{Mean} 
& 0.635/0.588
& 0.330/0.313
& 0.665/0.610
& 0.452/0.394
& \textbf{0.710/0.699} \\
\bottomrule
\end{tabular}

\begin{tablenotes}[flushleft]
\scriptsize
\item Entries are Spearman/Pearson correlations.
\item Positive values indicate increasing score with increasing severity.
\end{tablenotes}

\end{threeparttable}
\caption{Correlation between predicted severity and ground-truth degradation level for frozen VLM baselines and the proposed method under single and mixed degradation settings. (Test Set; patients : L506, L192, L310)}
\label{tab:severity_corr}
\end{table*}



\vspace{-1mm}
\paragraph{Benchmarked IQA metrics and VLMs.}
We first evaluate widely used full-reference CT image quality assessment (IQA) metrics, including PSNR, SSIM \cite{wang2004image}, VIF \cite{sheikh2006image}, LPIPS \cite{zhang2018unreasonable}, and DISTS \cite{ding2020image}, which are commonly used to assess CT enhancement methods \cite{kim2024systematic,clement2025ai,lei2023ct}. Their sensitivity to degradation severity is quantified using both \textbf{Spearman} rank correlation and \textbf{Pearson} correlation, together with severity-wise metric trends.
We also benchmark representative vision--language models, namely OpenCLIP \cite{cherti2023reproducible}, MedCLIP \cite{wang2022MedCLIP}, BioMedCLIP \cite{zhang2024bioMedCLIP}, and Merlin \cite{blankemeier2024merlin}. For each model, we measure the cosine embedding drift between reference and degraded CT images and report its Spearman and Pearson correlation with degradation severity.

Finally, we evaluate the proposed method, \textbf{SeSpeCT}, on two tasks: degradation classification and severity estimation. Classification performance is reported using Accuracy and F1 score, while severity prediction is evaluated using Mean Absolute Error (MAE), Root Mean Square Error (RMSE), Quadratic Weighted Kappa (QWK), and Spearman/Pearson correlation with ground-truth severity.


\subsection{Quantitative Results}

\subsubsection{Classical IQA metrics}


Table~\ref{tab:corr_metrics_degradations_full} and Table~\ref{tab:corr_metrics_degradations_testSet} summarize the correlation between degradation severity and widely used full-reference IQA metrics on the full dataset and the held-out test set, respectively. Detailed severity-wise statistics for the full dataset are provided in the supplementary material (Table~\ref{tab:fr_metrics_degradations_compact}). Overall, the results show that these metrics are not equally sensitive to all CT degradations. PSNR is most responsive to noise, whereas perceptual and structural metrics such as VIF and DISTS better capture blur severity. Structured degradations, including streaking and aliasing, exhibit clearer monotonic trends across most metrics.
\begin{table*}[h!]
\centering
\small
\setlength{\tabcolsep}{5pt}
\renewcommand{\arraystretch}{0.5}
\begin{tabular}{lccccc cc ccc}
\toprule
& \multicolumn{5}{c}{} & \multicolumn{2}{c}{Degradation classification} & \multicolumn{3}{c}{Severity estimation} \\
\cmidrule(lr){2-6} \cmidrule(lr){7-8} \cmidrule(lr){9-11}
Model & $B_{sem}$ & $B_{FFT}$ & $L_{reg}$ & $L_{rank}$ & $L_{con}$ & Acc$\uparrow$ & F1$\uparrow$ & MAE$\downarrow$ & RMSE$\downarrow$ & QWK$\uparrow$ \\
\midrule

w/o $B_{sem}$  & \ding{55} & \ding{51} & \ding{51} & \ding{51} & \ding{51} & 0.151 & 0.061 & 1.494 & 1.858 & 0.000 \\
w/o $B_{FFT}$  & \ding{51} & \ding{55} & \ding{51} & \ding{51} & \ding{51} & 0.698 & 0.684 & 0.696 & 0.855 & 0.683 \\

\midrule

w/o $L_{reg}$  & \ding{51} & \ding{51} & \ding{55} & \ding{51} & \ding{51} & 0.599 & 0.579 & 1.492 & 1.783 & 0.030 \\
w/o $L_{rank}$ & \ding{51} & \ding{51} & \ding{51} & \ding{55} & \ding{51} & 0.609 & 0.578 & 1.332 & 1.547 & 0.383 \\
w/o $L_{con}$  & \ding{51} & \ding{51} & \ding{51} & \ding{51} & \ding{55} & 0.695 & 0.686 & 0.966 & 1.173 & 0.571 \\

\midrule

\textbf{Proposed} & \ding{51} & \ding{51} & \ding{51} & \ding{51} & \ding{51} & \textbf{0.783} & \textbf{0.776} & \textbf{0.632} & \textbf{0.794} & \textbf{0.738} \\

\bottomrule
\end{tabular}

\vspace{2pt}
{\footnotesize
$B_{sem}$: semantic quality branch. \;
$B_{FFT}$: frequency branch. \;
$L_{reg}$: regression loss. \;
$L_{rank}$: ranking loss. \;
$L_{con}$: contrastive loss. \\
$\uparrow$ higher is better; $\downarrow$ lower is better.
}

\vspace{-3mm}

\caption{\textbf{Ablation study of SeSpeCT.} We remove one branch or loss term at a time and report degradation classification (Accuracy, F1) and severity estimation (MAE, RMSE, QWK) performance.}
\label{tab:ablation}
\end{table*}
In contrast, metal artifacts yield substantially weaker correlations, indicating that global quality metrics are less sensitive to spatially localized degradations. This is particularly important for CT enhancement evaluation, where commonly used metrics may fail to reflect restoration quality when degradations are localized rather than globally distributed.

\subsubsection{VLM embedding sensitivity to degradation}


Table~\ref{tab:severity_corr} reports the correlation between degradation severity and cosine embedding drift in frozen vision--language models on the test dataset. Detailed severity-wise drift statistics for the full dataset are provided in the supplementary material (Tables \ref{tab:corr_vlm_degradations} and \ref{tab:vlm_drift_degradations_compact}). Across degradations, OpenCLIP and BioMedCLIP show the strongest monotonic relationship between embedding drift and severity. In particular, blur, streaking, and aliasing produce consistent representation shifts that increase with degradation strength. 

By contrast, MedCLIP and Merlin show weaker severity correlations for several degradations, especially for metal artifacts and mixtures. More broadly, the VLM trends are consistent with those observed for classical IQA metrics: degradations that exhibit stronger monotonicity in structural or perceptual measures also tend to induce larger shifts in multimodal embedding space.

\begin{figure}[htbp]
\centering
\includegraphics[width=0.9\columnwidth]{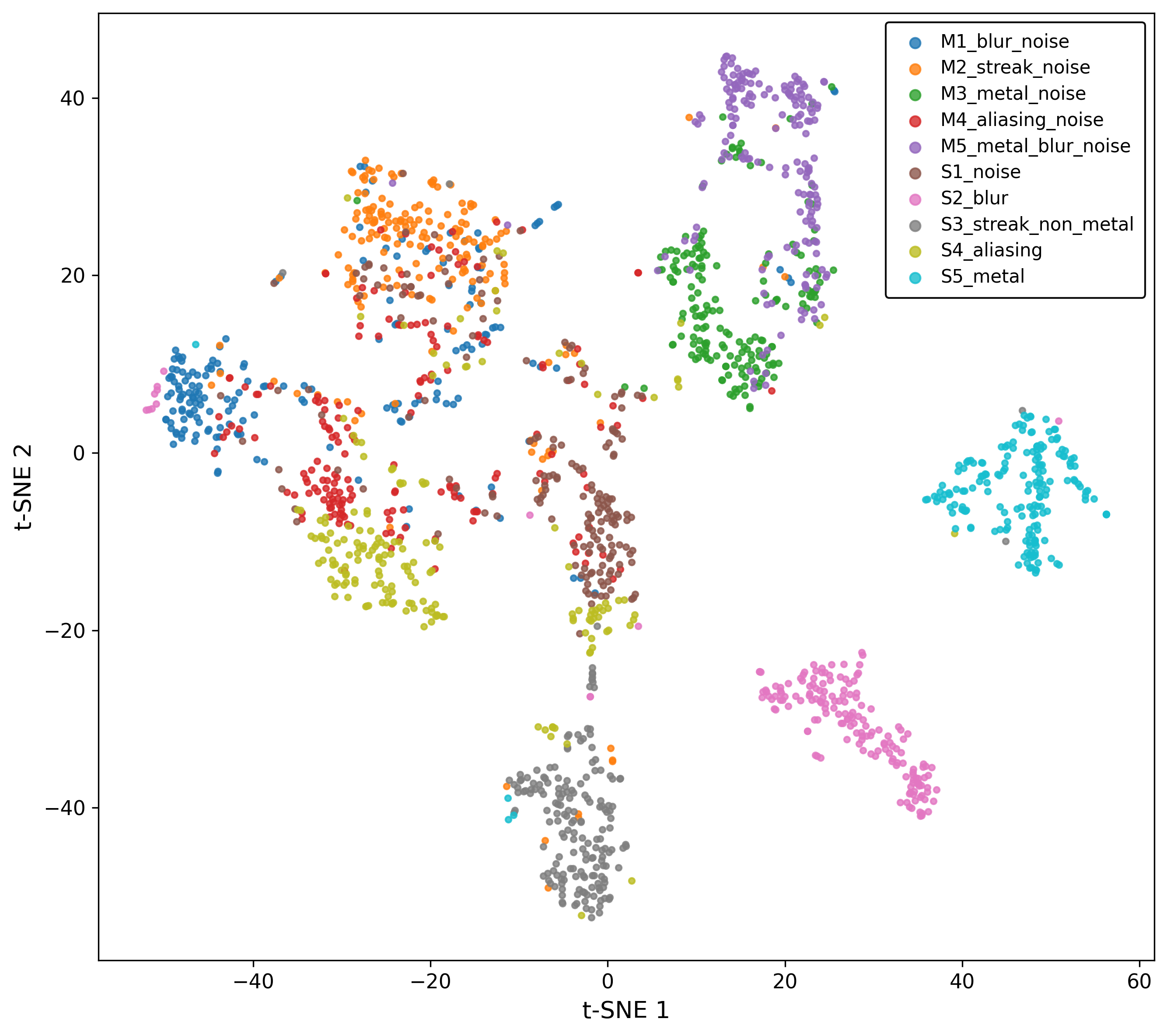}
\caption{\textbf{t-SNE visualization colored by degradation type.}
Samples form distinct clusters corresponding to different degradation categories, indicating that the learned representation captures degradation-specific characteristics.}
\label{fig:tsne_degradation}
\end{figure}
\subsubsection{Performance of SeSpeCT}
Table~\ref{tab:severity_corr} reports severity correlation on the held-out test set for SeSpeCT and frozen VLM baselines. In contrast to Table~\ref{tab:corr_vlm_degradations}, which analyzes representation drift on the full dataset, this experiment evaluates predictive severity estimation on unseen samples. SeSpeCT is trained for joint degradation analysis using the semantic--spectral representation described in Section~\ref{proposed_method}.

SeSpeCT achieves the highest correlation across both single and mixed settings, reaching a mean Spearman correlation of 0.780 for single degradations and 0.710 for mixtures, outperforming OpenCLIP (0.635) and BioMedCLIP (0.665). Gains are particularly pronounced for challenging cases such as metal artifacts, where correlations improve from $\rho < 0.12$ to $\rho = 0.435$.

These results indicate that combining semantic and spectral features improves sensitivity to both localized and structured degradations. Furthermore, Figure~\ref{fig:tsne_degradation} shows well-separated clusters across degradation types, confirming that the learned representation captures degradation-specific characteristics. Overall, SeSpeCT provides a more reliable and monotonic representation of degradation severity than frozen VLM features.

\subsubsection{Ablation study}

Table~\ref{tab:ablation} presents an ablation study of SeSpeCT. Removing the semantic branch causes the largest drop in performance, reducing classification accuracy from 0.783 to 0.151 and collapsing severity estimation performance, which highlights the importance of the prompt-derived semantic quality representation. Removing the spectral branch also leads to a substantial degradation, confirming that Fourier-domain cues provide complementary information beyond the semantic branch alone.
Among the loss terms, the regression loss is essential for meaningful severity prediction, while the ranking loss improves ordinal consistency across severity levels. The supervised contrastive term further improves both classification and severity estimation by structuring the fused embedding space.

\section{Conclusion}
\label{sec:conclusion}


This work addressed CT degradation analysis under a unified benchmark and method setting. We introduced \textbf{CT-DegradBench}, a physics-informed benchmark for controlled evaluation of CT degradation type and severity under both isolated and mixed settings. Our experiments showed that commonly used IQA metrics and frozen vision--language model embeddings do not consistently reflect degradation severity, particularly for localized and compound artifacts. We therefore proposed \textbf{SeSpeCT}, a semantic--spectral framework that combines a prompt-derived semantic quality axis from medical vision--language models with frequency-domain descriptors for joint degradation detection and severity estimation. Together, CT-DegradBench and SeSpeCT will support more reliable quality assessment and evaluation protocols for future CT restoration and analysis pipelines.

\
\section*{Acknowledgments}
This work was partially supported by the NIH grants R01-HL171376 and U01-CA268808.

{
\small

\bibliographystyle{ieeenat_fullname}
    \bibliography{main}
    \clearpage
}
\clearpage
\setcounter{page}{1}
\maketitlesupplementary
\section{Additional Quantitative Evaluation}

\subsection{Per-Degradation Classification Performance}

Table~\ref{tab:per_degradation_performance} reports classification accuracy and F1-score for each degradation across single distortions (S1–S5) and mixtures (M1–M5). Single degradations achieve consistently high performance, with most classes exceeding 0.98 accuracy. Blur (S2) and metal artifacts (S5) obtain the highest F1-scores (0.960 and 0.961), indicating that the proposed method effectively captures their distinctive spatial and spectral patterns.

In contrast, noise and aliasing exhibit lower F1-scores (0.587 and 0.252, respectively). This behavior likely reflects the difficulty of modeling aliasing artifacts, which arise from angular undersampling and are primarily expressed as high-frequency sampling patterns that are less effectively encoded by representations learned from semantic visual features. As expected, mixture degradations are generally more challenging than single distortions, since overlapping artifacts may partially obscure individual degradation cues. In particular, M4 (aliasing + noise) yields the lowest F1-score, suggesting increased ambiguity when frequency-domain artifacts interact with noise.
\begin{table}[htbp]
\centering
\scriptsize
\setlength{\tabcolsep}{4pt}
\begin{tabular}{lcc|lcc}
\hline
\multicolumn{3}{c|}{\textbf{Single Distortions}} & 
\multicolumn{3}{c}{\textbf{Mixture Distortions}} \\
\hline
Degradation & Accuracy $\uparrow$ & F1 $\uparrow$ & 
Degradation & Accuracy $\uparrow$ & F1 $\uparrow$ \\
\hline
S1\_noise      & 0.914 & 0.587 & M1\_b+n   & 0.935 & 0.598 \\
S2\_blur       & 0.992 & 0.960 & M2\_s+n   & 0.893 & 0.618 \\
S3\_streak\_nm & 0.982 & 0.911 & M3\_m+n   & 0.943 & 0.758 \\
S4\_aliasing   & 0.914 & 0.252 & M4\_a+n   & 0.876 & 0.438 \\
S5\_metal      & 0.992 & 0.961 & M5\_m+b+n & 0.955 & 0.757 \\
\hline
\end{tabular}

\vspace{2pt}
{\footnotesize
\textbf{Mixture shorthand:} b = blur, n = noise, s = streaks, a = aliasing, m = metal. \\
$\uparrow$ higher is better.
}

\caption{\textbf{Per-degradation classification performance of the proposed method.}
One-vs-rest accuracy and F1-score for single distortions (S1--S5) and mixtures (M1--M5).}
\label{tab:per_degradation_performance}
\end{table}
\begin{figure}[htbp]
    \centering
    \includegraphics[width=0.8\columnwidth]{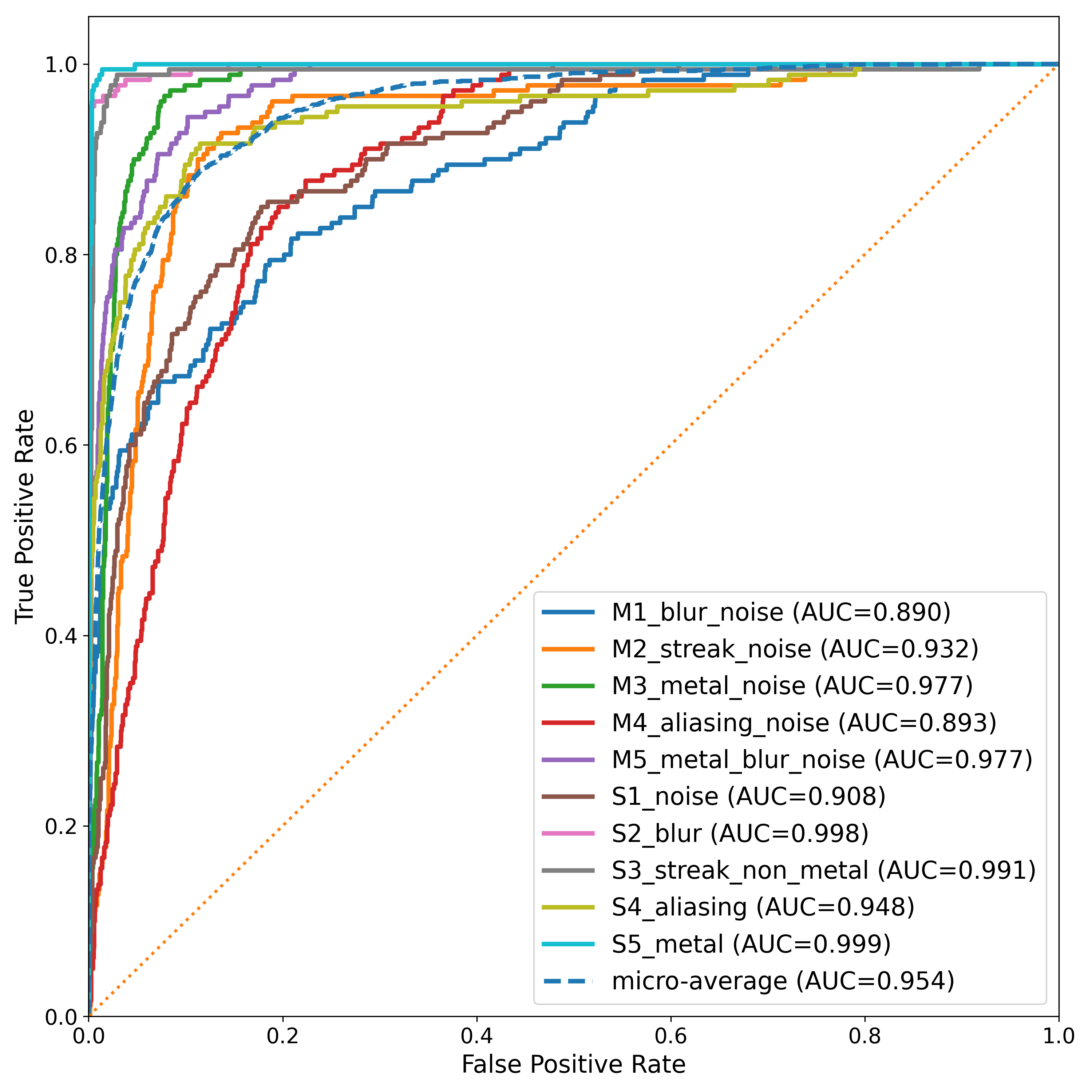}
    \caption{\textbf{ROC curves for degradation classification.}
One-vs-rest ROC curves are shown for each degradation category. The dashed line denotes the micro-average across all classes (AUC = 0.954).}    \label{fig:roc_degradation_classification}
\end{figure}
\subsection{ROC Analysis for Degradation Classification}

Figure~\ref{fig:roc_degradation_classification} presents ROC curves for all degradations. The micro-average AUC reaches 0.954, indicating strong overall degradations discrimination. Single degradations achieve near-perfect AUC values in several cases, confirming that the learned representation clearly separates artifact types. Although mixture degradations show slightly lower AUC values, their performance remains high, demonstrating that the model retains discriminative features even under combined artifacts.

\subsection{Severity Level Prediction Performance}
Table~\ref{tab:severity_per_degradation} reports severity prediction performance using MAE, RMSE, and Quadratic Weighted Kappa (QWK). For single degradations, blur and streak artifacts achieve the lowest errors.
\begin{table}[htbp]
\centering
\scriptsize
\setlength{\tabcolsep}{3pt}
\begin{tabular}{lccc|lccc}
\hline
\multicolumn{4}{c|}{\textbf{Single Distortions}} & 
\multicolumn{4}{c}{\textbf{Mixture Distortions}} \\
\hline
Degradation & MAE $\downarrow$ & RMSE $\downarrow$ & QWK $\uparrow$ &
Degradation & MAE $\downarrow$ & RMSE $\downarrow$ & QWK $\uparrow$ \\
\hline
S1\_noise      & 0.578 & 0.713 & 0.754 & M1\_b+n   & 0.821 & 1.030 & 0.656 \\
S2\_blur       & 0.387 & 0.567 & 0.869 & M2\_s+n   & 0.785 & 0.994 & 0.601 \\
S3\_streak\_nm & 0.495 & 0.626 & 0.803 & M3\_m+n   & 0.623 & 0.794 & 0.699 \\
S4\_aliasing   & 0.659 & 0.808 & 0.772 & M4\_a+n   & 0.820 & 1.009 & 0.566 \\
S5\_metal      & 0.989 & 1.242 & 0.396 & M5\_m+b+n & 0.599 & 0.771 & 0.716 \\
\hline
\end{tabular}

\vspace{2pt}
{\footnotesize
\textbf{Mixture shorthand:} b = blur, n = noise, s = streaks, a = aliasing, m = metal. \\
$\uparrow$ higher is better; $\downarrow$ lower is better.
}

\caption{\textbf{Per-degradation severity prediction performance.}
MAE, RMSE, and Quadratic Weighted Kappa (QWK) for severity estimation across single distortions (S1--S5) and mixtures (M1--M5).}
\label{tab:severity_per_degradation}
\end{table}
 Metal artifacts are more challenging to estimate, likely due to their dependence on implant geometry and position. Mixtures show moderate prediction errors but maintain reasonable agreement with ground truth, indicating that the model captures severity-related cues even in the presence of interacting degradations.

 \section{Embedding-Space Representation Analysis}
\subsection{t-SNE Visualization by Degradation Type}

Figure~\ref{fig:tsne_degradation} visualizes the learned embedding space projected using t-SNE and colored by degradation category. Distinct clusters emerge for most degradation types, indicating that the proposed representation encodes artifact-specific characteristics. Degradations that introduce strong structural patterns, such as blur and metal artifacts, form compact and well-separated clusters, suggesting that their spatial and spectral signatures are consistently captured by the model.

In contrast, degradations dominated by stochastic or high-frequency components, such as noise and aliasing, exhibit slightly more diffuse distributions. This behavior is expected, as these artifacts may share overlapping frequency characteristics with other degradations. For example, samples affected by aliasing occasionally appear closer to streak or noise clusters, reflecting the partial similarity between undersampling patterns and other frequency-domain distortions.

\subsection{t-SNE Visualization by Severity Level}

\begin{figure}[htbp]
\centering
\includegraphics[width=\columnwidth]{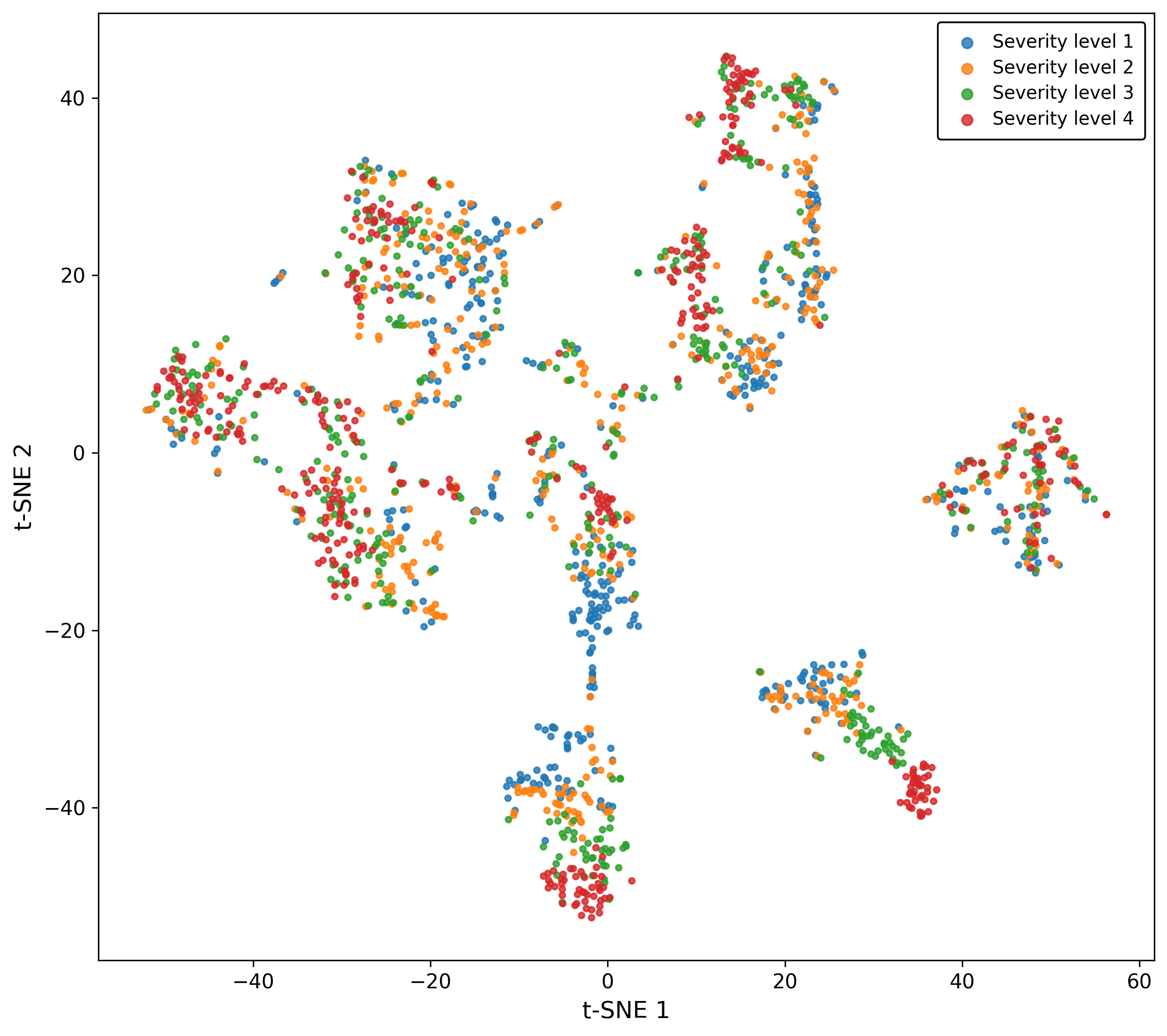}
\caption{\textbf{t-SNE visualization colored by degradation severity.}
The embedding reveals a gradual structure where samples are organized according to increasing degradation intensity.}
\label{fig:tsne_severity}
\end{figure}

Figure~\ref{fig:tsne_severity} shows the same embedding space colored by degradation severity. A gradual transition between severity levels can be observed within several clusters, indicating that the representation captures not only degradation type but also variations in degradation intensity. In particular, blur and streak artifacts display a clear progression from lower to higher severity levels along consistent directions in the embedding space.
entirely separate regions.
However, severity separation is less pronounced for noise and metal artifacts. This emphasizes the results presented previously in tables \ref{tab:per_degradation_performance} and \ref{tab:severity_per_degradation} for these degradations, where severity do not severely impact the visual characteristics. 

\subsection{t-SNE Visualization for Single and Mixed Degradations}

\begin{figure}[htbp]
\centering
\includegraphics[width=\columnwidth]{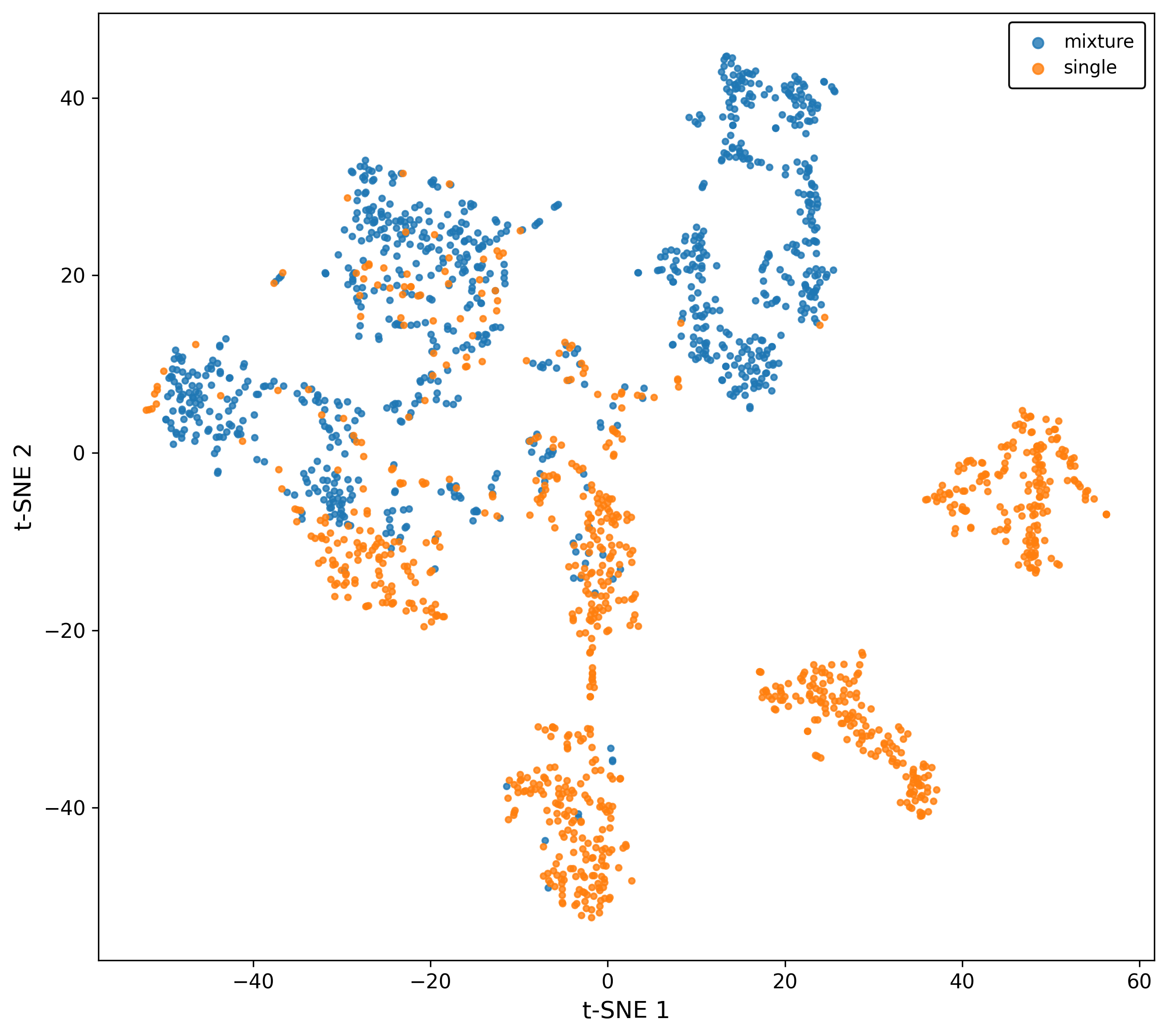}
\caption{\textbf{t-SNE visualization distinguishing single and mixed degradations.}
The representation separates samples containing a single degradation from those with multiple degradations, highlighting its ability to capture compositional corruption patterns.}
\label{fig:tsne_mixture}
\end{figure}

Figure~\ref{fig:tsne_mixture} highlights the distribution of single degradations versus mixtures. While many mixture samples remain close to the clusters associated with their dominant artifact type, they often appear near the boundaries between multiple clusters. This behavior suggests that mixed degradations combine characteristics from several artifact types, resulting in embeddings that lie between the corresponding single-degradation regions. For example, mixtures involving metal artifacts and noise tend to appear near the metal artifact cluster while exhibiting slight displacement toward regions associated with noise. These patterns indicate that the learned embedding space captures compositional degradation characteristics rather than assigning mixtures to 
\section{Extended Ablation Studies}
\subsection{Branch Contribution Analysis}

Table~\ref{tab:ablation} evaluates the contribution of each branch in the proposed semantic-spectral framework. Ablation results reveal that the two branches play complementary roles in capturing degradation characteristics.

Removing the semantic quality branch ($B_{\mathrm{sem}}$) causes a drastic drop in classification performance (Acc = 0.151, F1 = 0.061), indicating that the semantic representation learned from the vision--language model provides a crucial prior for distinguishing degradation types. Without this branch, the model relies primarily on frequency information, which alone shows insufficient performance to discriminate between artifacts.

In contrast, removing the frequency branch ($B_{\mathrm{FFT}}$) significantly reduces performance but to a lesser extent (Acc = 0.698, F1 = 0.684). This suggests that while semantic representations capture high-level structural cues, the spectral branch provides complementary information about high-frequency artifact patterns. Degradations such as aliasing, streak artifacts, and noise are particularly characterized by distinct frequency signatures, which explains the notable performance gain when the spectral branch is included.

\subsection{Loss Component Analysis}

The ablation results also highlight the importance of the different training objectives. Removing the regression loss ($L_{\mathrm{reg}}$) leads to a substantial degradation in severity estimation performance (RMSE increases from 0.794 to 1.783), confirming that explicit regression supervision is essential for learning a continuous severity representation.

Similarly, removing the ranking loss ($L_{\mathrm{rank}}$) noticeably affects the severity prediction metrics. The increase in RMSE and decrease in QWK suggest that the ranking constraint plays an important role in preserving the ordinal structure of degradation severity levels. In other words, it encourages the model to maintain consistent ordering between different severity levels, which is particularly important when the visual differences between adjacent levels are subtle.

Finally, removing the contrastive loss ($L_{\mathrm{con}}$) results in a decrease in both classification and severity estimation performance. This observation indicates that contrastive supervision helps structure the embedding space by pulling together samples with similar degradation characteristics while pushing apart dissimilar ones. 

Overall, the ablation study demonstrates that the semantic and spectral branches provide complementary information, while the combination of regression, ranking, and contrastive objectives encourages a representation space that is both discriminative and consistent with degradation severity.


\section{Additional Benchmark Details}
\label{sec:supp_dataset_details}

This section provides supplementary details on the prompt design used to construct the semantic quality axis, the physical motivation behind the degradation mixtures in CT-DegradBench, and the relation of CT-DegradBench to existing CT restoration datasets.

\subsection{Prompt design for the semantic quality axis}

Table~\ref{tab:supp_prompts_quality_axis} lists the high- and low-quality prompts used to construct the semantic quality axis in Section~\ref{proposed_method}. The high-quality prompts describe diagnostically reliable CT images with clear anatomy, sharp boundaries, and no visible artifacts, whereas the low-quality prompts describe degraded CT appearances corresponding to the main artifact families in CT-DegradBench: noise, blur, streaks, aliasing, and metal artifacts. These prompts define the semantic direction used by the semantic branch without task-specific fine-tuning.

\begin{table*}[t]
\centering
\small
\setlength{\tabcolsep}{4pt}
\renewcommand{\arraystretch}{1.15}

\begin{tabular*}{\textwidth}{@{\extracolsep{\fill}} p{0.48\textwidth} p{0.48\textwidth}}
\hline
\rowcolor{gray!15}
\textbf{High-Quality Prompts (H)} & \textbf{Low-Quality Prompts (L)} \\
\hline

\cellcolor{green!10}1. Axial abdominal CT slice with excellent diagnostic quality, sharp boundaries, clear organ detail, and no visible artifacts. &
\cellcolor{red!10}1. Abdominal CT slice with severe noise and grainy appearance that reduces visibility of anatomical structures. \\

\cellcolor{green!10}2. Diagnostic abdominal CT with clear anatomical structures, low noise, high contrast, and no streak artifacts. &
\cellcolor{red!10}2. Abdominal CT slice with strong blur and significant loss of sharpness. \\

\cellcolor{green!10}3. High-quality CT image with sharp edges, clean appearance, and good visibility of abdominal organs. &
\cellcolor{red!10}3. Abdominal CT slice with strong streak artifacts and reduced diagnostic quality. \\

\cellcolor{green!10}~ &
\cellcolor{red!10}4. Abdominal CT slice with sparse-view aliasing artifacts and distorted anatomical structures. \\

\cellcolor{green!10}~ &
\cellcolor{red!10}5. Abdominal CT slice with strong metal artifacts causing bright streaks and severe image corruption. \\
\hline
\end{tabular*}

\caption{\textbf{Prompt sets used to construct the semantic quality axis.} High-quality prompts describe artifact-free diagnostic CT images, while low-quality prompts represent common degradations including noise, blur, streak artifacts, aliasing, and metal artifacts.}
\label{tab:supp_prompts_quality_axis}
\end{table*}
\subsection{Clinical motivation for degradation mixtures}

Table~\ref{tab:mixtures_scenarios} summarizes the motivating scenarios and physical interpretation of the degradation mixtures used in CT-DegradBench. These mixtures are designed to reflect plausible CT acquisition conditions in which multiple degradation sources co-occur, particularly under low-dose imaging, sparse-view acquisition, or the presence of metallic implants.

\begin{table*}[htbp]
\centering
\scriptsize
\setlength{\tabcolsep}{2pt}
\begin{tabular}{ll}
\toprule
Mixture & Description \\
\midrule

\textbf{Loss of sharpness + Noise} 
& \textit{Scenario:} Low-dose CT with limited detector resolution. \\
& \textit{Mechanism:} Detector response (MTF) reduces spatial resolution (loss of sharpness) $\rightarrow$ low photon counts introduce noise. \\[2pt]

\textbf{Streaks + Noise} 
& \textit{Scenario:} Photon starvation in dense regions (e.g., shoulders or pelvis). \\
& \textit{Mechanism:} Highly attenuating paths produce streak artifacts $\rightarrow$ low-dose acquisition increases noise. \\[2pt]

\textbf{Metal + Noise} 
& \textit{Scenario:} CT scans acquired at low tube current with metallic implants (e.g., hip prosthesis). \\[2pt]
& \textit{Mechanism:} Photon starvation from metal produces streak artifacts $\rightarrow$ low photon counts introduce noise. \\

\textbf{Aliasing + Noise} 
& \textit{Scenario:} Accelerated or sparse-view low-dose CT acquisition. \\[2pt]
& \textit{Mechanism:} Angular undersampling produces aliasing $\rightarrow$ reduced photon counts introduce noise. \\

\textbf{Metal + Loss of sharpness + Noise} 
& \textit{Scenario:} CT imaging with metallic implants under low-dose conditions. \\
& \textit{Mechanism:} Metal beam hardening $\rightarrow$ detector response reduces spatial resolution $\rightarrow$ low photon counts introduce noise. \\

\bottomrule
\end{tabular}
\caption{Scenario motivation and physical interpretation of degradation mixtures in CT-DegradBench.}
\label{tab:mixtures_scenarios}
\end{table*}

\subsection{Comparison with existing CT datasets}

Table~\ref{tab:supp_dataset_comparison} compares CT-DegradBench with commonly used CT restoration datasets. Most existing datasets focus on a single degradation family, such as low-dose noise, sparse-view artifacts, or metal artifact reduction, and typically lack explicit severity control or mixed-degradation settings. In contrast, CT-DegradBench provides multiple degradation types, calibrated severity levels, and controlled mixtures within a unified benchmark.

\begin{table*}[htbp]
\centering
\scriptsize
\setlength{\tabcolsep}{2pt}

\begin{tabular}{l l l c c c}
\toprule
\textbf{Dataset} & \textbf{Task} & \textbf{Degradation Types} & \textbf{Severity per Degradation} & \textbf{Severity Control} & \textbf{Mixtures} \\
\midrule

\rowcolor{lowdose}
AAPM Mayo LDCT \cite{mccollough2016tu} & LDCT enhancement & Low-dose photon noise & Quarter dose level & \ding{55} & Implicit \\

\rowcolor{lowdose}
LDCT-and-Projection-data \cite{moen2021low} & LDCT enhancement & Low-dose degradations & Quarter dose level & \ding{55} & Implicit \\

\rowcolor{lowdose}
LoDoPaB-CT \cite{leuschner2021lodopab} & LDCT enhancement & Low-dose degradations & Quarter dose level & \ding{55} & Implicit \\

\rowcolor{lowdose}
Piglet LDCT Dataset \cite{yi2018sharpness} & LDCT enhancement & Low-dose degradations & Five dose levels & \ding{51} & Implicit \\
\midrule

\rowcolor{sparse}
AAPM Sparse-View CT Challenge \cite{sidky2022report} & Sparse-view reconstruction & Angular undersampling & Single level (128 views) & \ding{55} & \ding{55} \\

\midrule

\rowcolor{mar}
AAPM CT Metal Artifact Reduction Challenge \cite{haneda2025aapm} & Metal Artifact Reduction (MAR) & Simulated metal streak artifact & Not graded & \ding{55} & \ding{55} \\

\rowcolor{mar}
UCLH Stroke EIT Dataset~\cite{goren2017uclh} & Metal Artifact Reduction (MAR) & Metal-induced streak artifacts & Not explicitly graded & \ding{55} & \ding{55} \\

\midrule
\rowcolor{ours}
\textbf{CT-DegradBench (Ours)} & Degradation benchmarking & Noise, blur, streaks, aliasing, metal artifacts & \textbf{4 levels} & \ding{51} & \ding{51} \\

\bottomrule
\end{tabular}

\caption{Comparison with common CT datasets used in restoration and enhancement pipelines.}
\label{tab:supp_dataset_comparison}
\end{table*}
\begin{table*}[t]
\centering
\caption{Benchmarking results (mean $\pm$ std) for full-reference quality metrics across degradations and severity levels.}
\label{tab:fr_metrics_degradations_compact}
\scriptsize
\setlength{\tabcolsep}{3pt}
\renewcommand{\arraystretch}{1.03}

\begin{threeparttable}
\resizebox{\textwidth}{!}{%
\begin{tabular}{lccccc @{\hspace{0.9em}\vrule\hspace{0.9em}} lccccc}
\toprule
\rowcolor{gray!12}
\multicolumn{6}{c}{\textbf{Single distortions}} &
\multicolumn{6}{c}{\textbf{Mixtures}} \\
\cmidrule(lr){1-6}\cmidrule(lr){7-12}
\rowcolor{gray!12}
\textbf{Setting} & \textbf{PSNR} $\uparrow$ & \textbf{SSIM} $\uparrow$ & \textbf{VIF} $\uparrow$ & \textbf{LPIPS} $\downarrow$ & \textbf{DISTS} $\downarrow$ &
\textbf{Setting} & \textbf{PSNR} $\uparrow$ & \textbf{SSIM} $\uparrow$ & \textbf{VIF} $\uparrow$ & \textbf{LPIPS} $\downarrow$ & \textbf{DISTS} $\downarrow$ \\

\midrule

S1\_noise L0 & 29.67$\pm$1.95 & 0.8627$\pm$0.0482 & 0.3657$\pm$0.0877 & 0.3650$\pm$0.0730 & 0.1917$\pm$0.0505 &
M1\_b+n L0   & 28.20$\pm$2.31 & 0.7931$\pm$0.0909 & 0.2586$\pm$0.0615 & 0.4400$\pm$0.0698 & 0.2401$\pm$0.0482 \\

S1\_noise L1 & 27.14$\pm$2.44 & 0.7456$\pm$0.0978 & 0.2695$\pm$0.0787 & 0.4564$\pm$0.0730 & 0.2569$\pm$0.0484 &
M1\_b+n L1   & 27.54$\pm$2.56 & 0.7600$\pm$0.1089 & 0.2372$\pm$0.0647 & 0.4599$\pm$0.0741 & 0.2549$\pm$0.0503 \\

S1\_noise L2 & 25.96$\pm$2.68 & 0.6920$\pm$0.1142 & 0.2403$\pm$0.0739 & 0.4839$\pm$0.0710 & 0.2766$\pm$0.0453 &
M1\_b+n L2   & 25.56$\pm$2.85 & 0.6650$\pm$0.1273 & 0.1813$\pm$0.0557 & 0.5149$\pm$0.0634 & 0.2936$\pm$0.0400 \\

S1\_noise L3 & 23.08$\pm$3.10 & 0.5638$\pm$0.1373 & 0.1827$\pm$0.0631 & 0.5383$\pm$0.0657 & 0.3135$\pm$0.0384 &
M1\_b+n L3   & 24.40$\pm$2.77 & 0.6057$\pm$0.1224 & 0.1417$\pm$0.0380 & 0.5452$\pm$0.0520 & 0.3157$\pm$0.0298 \\

\midrule

S2\_blur L0  & 30.96$\pm$1.80 & 0.9260$\pm$0.0108 & 0.4629$\pm$0.0431 & 0.3539$\pm$0.0280 & 0.1363$\pm$0.0164 &
M2\_s+n L0   & 28.44$\pm$2.44 & 0.8027$\pm$0.0934 & 0.2973$\pm$0.0853 & 0.4254$\pm$0.0777 & 0.2319$\pm$0.0541 \\

S2\_blur L1  & 30.83$\pm$1.74 & 0.9225$\pm$0.0107 & 0.4264$\pm$0.0402 & 0.3787$\pm$0.0285 & 0.1525$\pm$0.0156 &
M2\_s+n L1   & 27.77$\pm$2.70 & 0.7720$\pm$0.1080 & 0.2769$\pm$0.0826 & 0.4453$\pm$0.0766 & 0.2455$\pm$0.0527 \\

S2\_blur L2  & 30.49$\pm$1.60 & 0.9115$\pm$0.0104 & 0.3534$\pm$0.0326 & 0.4128$\pm$0.0264 & 0.1881$\pm$0.0127 &
M2\_s+n L2   & 24.92$\pm$2.99 & 0.6505$\pm$0.1285 & 0.2109$\pm$0.0670 & 0.5094$\pm$0.0670 & 0.2881$\pm$0.0434 \\

S2\_blur L3  & 29.81$\pm$1.37 & 0.8846$\pm$0.0114 & 0.2541$\pm$0.0231 & 0.4600$\pm$0.0232 & 0.2370$\pm$0.0110 &
M2\_s+n L3   & 22.92$\pm$2.81 & 0.5642$\pm$0.1235 & 0.1744$\pm$0.0562 & 0.5442$\pm$0.0578 & 0.3119$\pm$0.0361 \\

\midrule

S3\_streak L0 & 31.58$\pm$1.69 & 0.9385$\pm$0.0082 & 0.5799$\pm$0.0444 & 0.2136$\pm$0.0145 & 0.0987$\pm$0.0130 &
M3\_m+n L0    & 23.89$\pm$0.90 & 0.7090$\pm$0.0622 & 0.2046$\pm$0.0406 & 0.4977$\pm$0.0406 & 0.2844$\pm$0.0284 \\

S3\_streak L1 & 31.28$\pm$1.59 & 0.9256$\pm$0.0094 & 0.5575$\pm$0.0439 & 0.2485$\pm$0.0165 & 0.1233$\pm$0.0165 &
M3\_m+n L1    & 23.05$\pm$1.45 & 0.6425$\pm$0.1087 & 0.1818$\pm$0.0503 & 0.5221$\pm$0.0537 & 0.3002$\pm$0.0351 \\

S3\_streak L2 & 30.29$\pm$1.31 & 0.8936$\pm$0.0166 & 0.5187$\pm$0.0439 & 0.2953$\pm$0.0180 & 0.1593$\pm$0.0197 &
M3\_m+n L2    & 20.89$\pm$1.66 & 0.4884$\pm$0.1067 & 0.1296$\pm$0.0357 & 0.5763$\pm$0.0426 & 0.3369$\pm$0.0233 \\

S3\_streak L3 & 27.83$\pm$0.87 & 0.8298$\pm$0.0305 & 0.4649$\pm$0.0440 & 0.3442$\pm$0.0199 & 0.1968$\pm$0.0207 &
M3\_m+n L3    & 19.19$\pm$1.13 & 0.3833$\pm$0.0694 & 0.0996$\pm$0.0226 & 0.6065$\pm$0.0322 & 0.3567$\pm$0.0158 \\

\midrule

S4\_alias L0 & 31.05$\pm$1.84 & 0.9234$\pm$0.0110 & 0.5232$\pm$0.0444 & 0.2343$\pm$0.0175 & 0.0923$\pm$0.0149 &
M4\_a+n L0   & 28.06$\pm$2.28 & 0.7750$\pm$0.0819 & 0.2729$\pm$0.0641 & 0.4513$\pm$0.0589 & 0.2552$\pm$0.0367 \\

S4\_alias L1 & 30.33$\pm$1.59 & 0.8647$\pm$0.0172 & 0.3936$\pm$0.0314 & 0.3641$\pm$0.0187 & 0.2069$\pm$0.0173 &
M4\_a+n L1   & 27.48$\pm$2.27 & 0.7439$\pm$0.0901 & 0.2543$\pm$0.0638 & 0.4680$\pm$0.0589 & 0.2663$\pm$0.0357 \\

S4\_alias L2 & 29.40$\pm$1.33 & 0.7896$\pm$0.0268 & 0.3076$\pm$0.0244 & 0.4322$\pm$0.0157 & 0.2538$\pm$0.0127 &
M4\_a+n L2   & 25.25$\pm$2.60 & 0.6134$\pm$0.1036 & 0.1885$\pm$0.0523 & 0.5311$\pm$0.0515 & 0.3018$\pm$0.0271 \\

S4\_alias L3 & 28.57$\pm$1.16 & 0.7267$\pm$0.0331 & 0.2546$\pm$0.0202 & 0.4697$\pm$0.0138 & 0.2793$\pm$0.0101 &
M4\_a+n L3   & 23.87$\pm$2.79 & 0.5478$\pm$0.1020 & 0.1606$\pm$0.0409 & 0.5560$\pm$0.0485 & 0.3149$\pm$0.0258 \\

\midrule

S5\_metal L0 & 25.28$\pm$0.86 & 0.8527$\pm$0.0197 & 0.3780$\pm$0.0397 & 0.3940$\pm$0.0291 & 0.2004$\pm$0.0179 &
M5\_m+b+n L0 & 24.11$\pm$0.94 & 0.7314$\pm$0.0676 & 0.1960$\pm$0.0385 & 0.4890$\pm$0.0425 & 0.2777$\pm$0.0293 \\

S5\_metal L1 & 24.61$\pm$0.80 & 0.8414$\pm$0.0204 & 0.3686$\pm$0.0385 & 0.3946$\pm$0.0279 & 0.2058$\pm$0.0177 &
M5\_m+b+n L1 & 23.11$\pm$1.34 & 0.6469$\pm$0.1022 & 0.1638$\pm$0.0420 & 0.5262$\pm$0.0491 & 0.3023$\pm$0.0317 \\

S5\_metal L2 & 24.02$\pm$0.74 & 0.8311$\pm$0.0213 & 0.3597$\pm$0.0373 & 0.3960$\pm$0.0273 & 0.2109$\pm$0.0177 &
M5\_m+b+n L2 & 21.13$\pm$1.52 & 0.4979$\pm$0.1023 & 0.1128$\pm$0.0306 & 0.5802$\pm$0.0387 & 0.3391$\pm$0.0217 \\

S5\_metal L3 & 23.58$\pm$0.69 & 0.8221$\pm$0.0222 & 0.3519$\pm$0.0363 & 0.3976$\pm$0.0265 & 0.2144$\pm$0.0174 &
M5\_m+b+n L3 & 19.52$\pm$1.44 & 0.3942$\pm$0.0918 & 0.0804$\pm$0.0195 & 0.6112$\pm$0.0340 & 0.3595$\pm$0.0187 \\

\bottomrule
\end{tabular}%
}

\begin{tablenotes}[flushleft]
\footnotesize

\item \textbf{Setting keywords:} S1-S5 = single degradations; M1-M5 = mixtures; L0-L3 = severity levels (increasing with $L$).
\item \textbf{Mixture shorthand:} b = blur, n = noise, s = streaks, a = aliasing, m = metal.
\item $\uparrow$ higher is better; $\downarrow$ lower is better.
\end{tablenotes}
\end{threeparttable}
\end{table*}

\begin{table*}[h!]
\centering


\scriptsize
\setlength{\tabcolsep}{6pt}
\renewcommand{\arraystretch}{0.5}
\begin{threeparttable}
\resizebox{\textwidth}{!}{%
\begin{tabular}{lcccc @{\hspace{0.6em}\vrule\hspace{0.6em}} lcccc}
\toprule
\rowcolor{gray!12}
\multicolumn{5}{c}{\textbf{Single distortions} ($\rho$ / $r$)} &
\multicolumn{5}{c}{\textbf{Mixtures} ($\rho$ / $r$)} \\
\cmidrule(lr){1-5}\cmidrule(lr){6-10}
\rowcolor{gray!12}
\textbf{Setting} &
\textbf{OpenCLIP \cite{cherti2023reproducible}} &
\textbf{MedCLIP \cite{wang2022MedCLIP}} &
\textbf{BioMedCLIP \cite{zhang2024bioMedCLIP}} &
\textbf{Merlin \cite{blankemeier2024merlin}} &
\textbf{Setting} &
\textbf{OpenCLIP \cite{cherti2023reproducible}} &
\textbf{MedCLIP \cite{wang2022MedCLIP}} &
\textbf{BioMedCLIP \cite{zhang2024bioMedCLIP}} &
\textbf{Merlin \cite{blankemeier2024merlin}} \\
\midrule

S1\_noise
& 0.6414/0.6422 & 0.3253/0.2931 & 0.6401/0.6046 & 0.6219/0.5272 &
M1\_b+n
& 0.5453/0.5557 & 0.2470/0.2269 & 0.5963/0.5807 & 0.4139/0.3512 \\

S2\_blur
& 0.5395/0.5261 & 0.5527/0.5128 & 0.8854/0.7817 & 0.1712/0.1679 &
M2\_s+n
& 0.6134/0.6120 & 0.2706/0.2622 & 0.6823/0.6613 & 0.5126/0.4367 \\

S3\_streak
& 0.7734/0.6647 & 0.5770/0.5311 & 0.8962/0.8428 & 0.2818/0.2862 &
M3\_m+n
& 0.7462/0.7448 & 0.1985/0.1941 & 0.6272/0.6088 & 0.6717/0.5891 \\

S4\_aliasing
& 0.8876/0.8419 & 0.7371/0.6268 & 0.9115/0.8251 & 0.5651/0.5180 &
M4\_a+n
& 0.7148/0.6939 & 0.2230/0.2113 & 0.6473/0.6036 & 0.4597/0.4152 \\

S5\_metal
& 0.2864/0.2845 & 0.0778/0.0688 & 0.0989/0.0957 & 0.0645/0.0640 &
M5\_m+b+n
& 0.7411/0.7438 & 0.1906/0.1985 & 0.6569/0.6402 & 0.6679/0.5830 \\

\midrule
\textbf{Mean}
& 0.6257/0.5919 & 0.4540/0.4065 & \textbf{0.6864/0.6300} & 0.3409/0.3127
&
\textbf{Mean}
& \textbf{0.6722/0.6700} & 0.2259/0.2186 & 0.6420/0.6189 & 0.5452/0.4750 \\

\bottomrule
\end{tabular}%
}

\begin{tablenotes}[flushleft]
\footnotesize
\item Entries are reported as \textbf{$\rho$/$r$} (Spearman/Pearson).
\item \textbf{Setting keywords:} S1-S5 = single degradations; M1-M5 = mixtures.
\end{tablenotes}
\end{threeparttable}
\caption{Correlation between VLM embedding drift and degradation severity for single degradations and degradation mixtures. Entries are reported as Spearman/Pearson correlations.  (Full Dataset)}
\label{tab:corr_vlm_degradations}
\end{table*}

\begin{table*}[t]
\centering
\caption{Benchmarking results (mean $\pm$ std) for VLM embedding drift across degradations and severity levels).}
\label{tab:vlm_drift_degradations_compact}
\centering
\scriptsize
\setlength{\tabcolsep}{3pt}
\renewcommand{\arraystretch}{1.05}
\scriptsize
\setlength{\tabcolsep}{3pt}
\renewcommand{\arraystretch}{1.03}

\begin{threeparttable}
\resizebox{\textwidth}{!}{%
\begin{tabular}{lccccc @{\hspace{0.9em}\vrule\hspace{0.9em}} lccccc}
\toprule
\rowcolor{gray!12}
\multicolumn{5}{c}{\textbf{Single distortions}} &
\multicolumn{5}{c}{\textbf{Mixtures}} \\
\cmidrule(lr){1-5}\cmidrule(lr){6-10}
\rowcolor{gray!12}
\textbf{Setting} &
\textbf{OpenCLIP \cite{cherti2023reproducible}} &
\textbf{MedCLIP \cite{wang2022MedCLIP}} &
\textbf{BioMedCLIP \cite{zhang2024bioMedCLIP}} &
\textbf{Merlin \cite{blankemeier2024merlin}} &
\textbf{Setting} &
\textbf{OpenCLIP \cite{cherti2023reproducible}} &
\textbf{MedCLIP \cite{wang2022MedCLIP}} &
\textbf{BioMedCLIP \cite{zhang2024bioMedCLIP}} &
\textbf{Merlin \cite{blankemeier2024merlin}} \\
\midrule

S1\_noise L0 & 0.0636$\pm$0.0255 & 0.0537$\pm$0.0471 & 0.0150$\pm$0.0121 & 0.0021$\pm$0.0024 &
M1\_b+n L0   & 0.0885$\pm$0.0350 & 0.0732$\pm$0.0479 & 0.0475$\pm$0.0332 & 0.0049$\pm$0.0066 \\

S1\_noise L1 & 0.0968$\pm$0.0374 & 0.0767$\pm$0.0523 & 0.0503$\pm$0.0390 & 0.0069$\pm$0.0070 &
M1\_b+n L1   & 0.1003$\pm$0.0415 & 0.0780$\pm$0.0497 & 0.0614$\pm$0.0474 & 0.0058$\pm$0.0065 \\

S1\_noise L2 & 0.1147$\pm$0.0420 & 0.0844$\pm$0.0543 & 0.0697$\pm$0.0510 & 0.0102$\pm$0.0096 &
M1\_b+n L2   & 0.1382$\pm$0.0556 & 0.0933$\pm$0.0545 & 0.1087$\pm$0.0662 & 0.0111$\pm$0.0108 \\

S1\_noise L3 & 0.1722$\pm$0.0660 & 0.0999$\pm$0.0592 & 0.1258$\pm$0.0793 & 0.0205$\pm$0.0171 &
M1\_b+n L3   & 0.1752$\pm$0.0613 & 0.1046$\pm$0.0580 & 0.1548$\pm$0.0736 & 0.0141$\pm$0.0137 \\

\midrule

S2\_blur L0 & 0.0492$\pm$0.0169 & 0.0329$\pm$0.0199 & 0.0100$\pm$0.0056 & 0.0007$\pm$0.0005 &
M2\_s+n L0  & 0.0837$\pm$0.0336 & 0.0700$\pm$0.0490 & 0.0491$\pm$0.0319 & 0.0041$\pm$0.0048 \\

S2\_blur L1 & 0.0519$\pm$0.0180 & 0.0395$\pm$0.0236 & 0.0144$\pm$0.0072 & 0.0007$\pm$0.0005 &
M2\_s+n L1  & 0.0925$\pm$0.0373 & 0.0769$\pm$0.0514 & 0.0666$\pm$0.0488 & 0.0057$\pm$0.0076 \\

S2\_blur L2 & 0.0601$\pm$0.0211 & 0.0567$\pm$0.0324 & 0.0307$\pm$0.0131 & 0.0007$\pm$0.0005 &
M2\_s+n L2  & 0.1405$\pm$0.0551 & 0.1002$\pm$0.0654 & 0.1297$\pm$0.0681 & 0.0125$\pm$0.0145 \\

S2\_blur L3 & 0.0872$\pm$0.0273 & 0.0796$\pm$0.0375 & 0.0729$\pm$0.0269 & 0.0009$\pm$0.0006 &
M2\_s+n L3  & 0.1814$\pm$0.0620 & 0.1089$\pm$0.0630 & 0.1736$\pm$0.0624 & 0.0181$\pm$0.0147 \\

\midrule

S3\_streak L0 & 0.0366$\pm$0.0140 & 0.0183$\pm$0.0127 & 0.0084$\pm$0.0066 & 0.0004$\pm$0.0003 &
M3\_m+n L0    & 0.1884$\pm$0.0572 & 0.1046$\pm$0.0569 & 0.2022$\pm$0.0494 & 0.0052$\pm$0.0050 \\

S3\_streak L1 & 0.0451$\pm$0.0163 & 0.0296$\pm$0.0179 & 0.0331$\pm$0.0244 & 0.0004$\pm$0.0003 &
M3\_m+n L1    & 0.2331$\pm$0.0908 & 0.1110$\pm$0.0593 & 0.2237$\pm$0.0505 & 0.0088$\pm$0.0083 \\

S3\_streak L2 & 0.0629$\pm$0.0253 & 0.0453$\pm$0.0272 & 0.0927$\pm$0.0415 & 0.0005$\pm$0.0004 &
M3\_m+n L2    & 0.3337$\pm$0.0766 & 0.1271$\pm$0.0677 & 0.2657$\pm$0.0536 & 0.0185$\pm$0.0150 \\

S3\_streak L3 & 0.1029$\pm$0.0403 & 0.0580$\pm$0.0332 & 0.1568$\pm$0.0492 & 0.0007$\pm$0.0005 &
M3\_m+n L3    & 0.3924$\pm$0.0508 & 0.1366$\pm$0.0692 & 0.3010$\pm$0.0430 & 0.0279$\pm$0.0155 \\

\midrule

S4\_alias L0 & 0.0411$\pm$0.0150 & 0.0123$\pm$0.0083 & 0.0044$\pm$0.0037 & 0.0006$\pm$0.0005 &
M4\_a+n L0   & 0.0953$\pm$0.0305 & 0.0749$\pm$0.0496 & 0.0443$\pm$0.0318 & 0.0048$\pm$0.0054 \\

S4\_alias L1 & 0.0843$\pm$0.0282 & 0.0504$\pm$0.0295 & 0.0242$\pm$0.0143 & 0.0009$\pm$0.0006 &
M4\_a+n L1   & 0.1092$\pm$0.0394 & 0.0784$\pm$0.0480 & 0.0564$\pm$0.0399 & 0.0058$\pm$0.0061 \\

S4\_alias L2 & 0.1410$\pm$0.0411 & 0.0710$\pm$0.0376 & 0.0542$\pm$0.0215 & 0.0015$\pm$0.0009 &
M4\_a+n L2   & 0.1705$\pm$0.0486 & 0.0969$\pm$0.0563 & 0.1113$\pm$0.0588 & 0.0115$\pm$0.0104 \\

S4\_alias L3 & 0.1763$\pm$0.0403 & 0.0847$\pm$0.0434 & 0.0926$\pm$0.0358 & 0.0020$\pm$0.0012 &
M4\_a+n L3   & 0.1993$\pm$0.0483 & 0.1031$\pm$0.0585 & 0.1437$\pm$0.0676 & 0.0170$\pm$0.0160 \\

\midrule

S5\_metal L0 & 0.0955$\pm$0.0229 & 0.1069$\pm$0.0498 & 0.1860$\pm$0.0515 & 0.0011$\pm$0.0007 &
M5\_m+b+n L0 & 0.1745$\pm$0.0618 & 0.1051$\pm$0.0527 & 0.2013$\pm$0.0484 & 0.0040$\pm$0.0043 \\

S5\_metal L1 & 0.1022$\pm$0.0230 & 0.1106$\pm$0.0490 & 0.1952$\pm$0.0518 & 0.0012$\pm$0.0008 &
M5\_m+b+n L1 & 0.2301$\pm$0.0848 & 0.1137$\pm$0.0577 & 0.2225$\pm$0.0525 & 0.0080$\pm$0.0073 \\

S5\_metal L2 & 0.1088$\pm$0.0233 & 0.1140$\pm$0.0488 & 0.1977$\pm$0.0520 & 0.0012$\pm$0.0008 &
M5\_m+b+n L2 & 0.3264$\pm$0.0750 & 0.1272$\pm$0.0661 & 0.2666$\pm$0.0468 & 0.0168$\pm$0.0125 \\

S5\_metal L3 & 0.1140$\pm$0.0244 & 0.1158$\pm$0.0473 & 0.2000$\pm$0.0523 & 0.0013$\pm$0.0008 &
M5\_m+b+n L3 & 0.3783$\pm$0.0579 & 0.1376$\pm$0.0679 & 0.3068$\pm$0.0448 & 0.0246$\pm$0.0158 \\

\bottomrule
\end{tabular}%
}
\begin{tablenotes}[flushleft]
\footnotesize
\item 
\textbf{Setting keywords:} S1-S5 = single degradations; M1-M5 = mixtures; L0-L3 = severity levels. 

\end{tablenotes}
\end{threeparttable}
\end{table*}

\end{document}